\documentclass{article} 
\usepackage{iclr2027_conference,times}

\usepackage{amsmath,amsfonts,bm}

\def\eqref#1{equation~\ref{#1}}

\def\1{\bm{1}}

\DeclareMathAlphabet{\mathsfit}{\encodingdefault}{\sfdefault}{m}{sl}
\SetMathAlphabet{\mathsfit}{bold}{\encodingdefault}{\sfdefault}{bx}{n}

\usepackage{hyperref}
\usepackage{url}
\usepackage{graphicx}
\usepackage{cleveref}
\usepackage{booktabs}
\usepackage{amsthm}
\usepackage{fontawesome5} 

\crefname{appendix}{Appendix}{Appendices}
\Crefname{appendix}{Appendix}{Appendices} 

\title{Beyond the Training Horizon: Mechanisms and Limits of Length Generalization in Looped Transformers}

\author{
\parbox{\textwidth}{\centering \normalfont
\textbf{Jia Liang}$^{2}$ \qquad
\textbf{Xi Jin}$^{1}$ \qquad
\textbf{Liangming Pan}$^{1}$$\thanks{Corresponding author.} $ \\[0.5ex]
$^{1}$MOE Key Laboratory of Computational Linguistics, \\ School of Computer Science, Peking University \\[0.5ex]
$^{2}$Institute for Computational and Mathematical Engineering, Stanford University  \\[0.5ex]
\texttt{jialiang2015@gmail.com} \qquad \texttt{jinxi26@stu.pku.edu.cn} \\  
\texttt{liangmingpan@pku.edu.cn}  \\ \href{https://github.com/jialiang19/looped-transformer-mechanism-lengthGen}{\textcolor{black}{\faGithub} \ \texttt{Looped-Transformer-Mechanism-LengthGen}} 
}
}

\iclrfinalcopy 

\begin{document}

\maketitle
\lhead{Preprint. Under review}

\begin{abstract}
Looped Transformers can generalize to reasoning chains longer than those encountered during training, but the computations enabling this behavior and limiting its extent remain unclear. We mechanistically compare two looped-Transformer configurations, which we call the Matched-Recurrence Looped Transformer (MR-Loop) and Decoupled-Recurrence Looped Transformer (DR-Loop), reflecting their respective recurrence-training schemes. We evaluate polynomial iteration, finite-state composition, and knowledge-graph traversal using detailed mechanistic analysis. Attention analysis, intermediate-state decoding, and causal interventions reveal distinct mechanisms learned under final-answer supervision. MR-Loop updates an intermediate state at a fixed readout while advancing relation selection through adjacent-token interactions and a transferable progress cue. DR-Loop instead propagates intermediate states across relation positions, forming an advancing computational frontier. However, both mechanisms become unreliable at greater depths: MR-Loop exhibits degradation of its readout state and progress cues, while DR-Loop exhibits declining reliability of state propagation. Limited self-correction allows local errors to persist and compound. Across both models, we uncover a common representational principle: recurrent states encode not only task-relevant content but also its computational status—whether that content remains in a form that can support subsequent computation. Transferable live–consumed and fresh–aged residual directions causally control whether represented information can participate in subsequent computation, including beyond the training horizon. We further show that length generalization need not rely on faithful step-by-step reasoning, as Looped Transformers can exploit task structure without explicitly representing every intermediate state. 
\end{abstract}

\section{Introduction}






\label{sec:introduction}

Many reasoning problems in LLMs require executing the same computation for a variable and previously unseen number of steps \citep{schwarzschild2021algorithm,thomm2024limits}. This motivates \textbf{length generalization}: whether a model trained on shorter computational chains can solve longer instances governed by the same iterative rule \citep{anil2022exploring,huang2025formal}. Standard Transformers often struggle in this regime because the required number of sequential operations grows while their computational depth remains fixed \citep{hahn2020theoretical,zhou2024robust}. \textbf{Looped Transformers} offer a natural alternative: by repeatedly applying shared Transformer blocks, they can allocate additional computation to longer instances \citep{dehghani2019universal,geiping2025scaling}.

Recent works show that this recurrent structure can support length generalization \citep{xu2025expressive,labovich2026stability}. \citet{fan2024looped} demonstrate extrapolation to unseen input lengths on iterative algorithmic tasks, while \citet{kohli2026loop} show that increasing recurrent depth at inference time can enable deeper implicit compositions. Yet this extrapolation is bounded: performance can deteriorate at sufficiently long horizons, and excessive recurrence can even alter otherwise correct predictions \citep{kohli2026loop,yang2026stabilizing}. These observations raise a mechanistic question that behavioral accuracy alone cannot answer: \emph{what computation is reused beyond the training horizon, and what causes that computation to eventually break down?} \citep{blayney2026mechanistic}

Answering this requires understanding how recurrent computation is internally organized: \emph{where} intermediate states are represented, \emph{how} they are updated across recurrent steps, and \emph{which} computations are reused at unseen recurrent depths \citep{saunshi2025latent,fan2026bridging}. It is also unclear whether different looped architectures realize length generalization through a common mechanism or through mechanistically distinct recurrent algorithms \citep{giannou2023looped,yang2024looped}. Finally, successful extrapolation does not by itself establish that the model faithfully executes the intended step-by-step computation \citep{turpin2023unfaithful,lanham2023faithfulness}.

In this paper, we compare two existing looped-Transformer configurations from \citet{fan2024looped} and \citet{kohli2026loop}, which we refer to as \textit{Matched-Recurrence Looped Transformer (MR-Loop)} and \textit{Decoupled-Recurrence Looped Transformer (DR-Loop)} to highlight a salient difference in how recurrent depth is coupled to instance complexity. These labels are descriptive shorthand for the complete model–training configurations, not a controlled recurrence ablation; other architectural differences are detailed in \Cref{sec:loop_transformers} and \Cref{app:architectures}.

We evaluate both models on three controlled iterative task families: polynomial iteration, finite-state function composition, and multi-hop knowledge-graph traversal. These tasks require additional applications of a known transition rule as the problem horizon grows, while ground-truth intermediate states remain available for analysis. Models receive supervision only on the final answer, allowing us to examine the learned recurrent computation without explicit step-level supervision. Across task families, we vary task-specific structural parameters and combine attention analysis, intermediate-state decoding, and causal interventions to study the following three key questions:
\textbf{RQ1 (Mechanism):} \textit{How do looped transformers achieve length generalization?}
\textbf{RQ2 (Limits):} \textit{Why does length generalization fail at extremely long horizons?}
\textbf{RQ3 (Faithfulness):} \textit{Does successful length generalization imply faithful iterative computation?} Together, these questions move beyond whether additional recurrence improves accuracy: they ask when recurrent computation forms a reusable algorithm, what internal state must be preserved for continued execution, and what limits its extension to longer horizons---providing design principles for recurrent reasoning systems that can reliably extend, rather than merely increase, their test-time computation.

Our results reveal three main findings. First, MR-Loop and DR-Loop achieve length generalization through distinct recurrent mechanisms: MR-Loop maintains the intermediate state at a fixed readout position while advancing relation selection, whereas DR-Loop propagates intermediate states through an advancing computational frontier. In both cases, successful reuse depends not only on semantic content but also on its \emph{computational status}, captured by two pairs of transferable residual directions: live vs.\ consumed, and fresh vs.\ aged. Second, extreme-length failure arises from two sources: degradation of the states that support recurrent computation, and limited self-correction. MR-Loop's recurrent readout and progress signals deteriorate, while DR-Loop's propagated intermediate states become less reliable, which allows local errors to persist and compound over longer trajectories. Third, length generalization is task-dependent and does not imply faithful step-by-step reasoning; both models can exploit shortcut structure and reach correct extrapolated answers without consistently representing the intended intermediate trajectory. Together, these findings distinguish \emph{learning a reusable recurrent computation} from \emph{faithfully executing it} beyond the training horizon.

\section{Related Work}
\label{sec:related_work}

Looped Transformers repeatedly apply shared Transformer blocks, allowing computational depth to grow without adding parameters. Building on recurrent-depth architectures with adaptive computation \citep{dehghani2019universal}, subsequent work has established the potential of looped Transformers through constructive simulation of iterative algorithms \citep{giannou2023looped}, inductive biases for learning iterative updates \citep{yang2024looped}, and theoretical expressivity results \citep{xu2025expressive}. A natural question is whether this additional recurrent computation enables models to solve problems requiring more computational steps than encountered during training. While length generalization has been extensively studied in standard feedforward Transformers \citep{zhou2024algorithms,zhou2024robust,cho2024position,sabbaghi2024structural}, recent work has developed training schemes that enable looped Transformers to generalize to longer computational horizons \citep{fan2024looped,kohli2026loop}. Recent analyses have also examined looped Transformers mechanistically: \citet{blayney2026mechanistic} study layer-specific fixed-point behavior in cyclically recurrent models, whereas \citet{labovich2026stability} analyzes how fixed-point geometry, input dependence, recall, and normalization affect stable extrapolation, primarily characterizing recurrent dynamics. What remains unclear is the task-level mechanism by which looped Transformers realize---and eventually lose---length generalization. Our work addresses this gap by asking how they achieve length generalization, why extrapolation fails at extreme horizons, and whether successful extrapolation implies faithful iterative computation.

\section{Preliminaries}

We study recurrent computation in looped Transformers on sequential
composition tasks. We distinguish two notions of computational depth.
The \emph{task horizon} $H$ is the number of ground-truth transitions
required by an example, whereas the \emph{recurrent depth} $K$ is the
number of times the shared Transformer stack is applied. These quantities need not coincide: one recurrent pass updates all sequence positions simultaneously and does not in general correspond to one task transition.

\subsection{Looped Transformers}
\label{sec:loop_transformers}

Let $\mathbf{x}=(x_0,\ldots,x_{m-1})$ be an input sequence,
$\mathbf{e}=E(\mathbf{x})$ its token embeddings, and
$B_\theta$ a stack of $L$ causal Transformer blocks. A looped
Transformer repeatedly applies the same parameterized stack,
\[
    \mathbf{h}^{(t+1)}
    =
    B_\theta\!\left(
        \Phi(\mathbf{h}^{(t)},\mathbf{e})
    \right),
    \qquad t=0,\ldots,K-1,
\]
where $\Phi$ specifies how the original input embeddings enter the
recurrent computation. Thus, increasing $K$ increases computational
depth without introducing new Transformer parameters.

We study two existing instantiations of this framework. In the
\emph{Matched-Recurrence Looped Transformer} (MR-Loop;
\citealp{fan2024looped}), the input embeddings are reinjected at every
pass,
\[
    \mathbf{h}^{(0)}=0,\qquad
    \mathbf{h}^{(t+1)}
    = B_\theta(\mathbf{h}^{(t)}+\mathbf{e}),
\]
and the prescribed recurrence depth is coupled to input length
($K=T=n$ in our notation). The answer is read from an \texttt{[EQ]}
delimiter.

In the \emph{Decoupled-Recurrence Looped Transformer} (DR-Loop;
\citealp{kohli2026loop}), the input is embedded once,
\[
    \mathbf{h}^{(0)}=\mathbf{e},\qquad
    \mathbf{h}^{(t+1)}
    = B_\theta(\mathbf{h}^{(t)}),
\]
while recurrent depth is sampled independently of input length during
training and can be chosen separately at inference time. The answer is
read from a padded output position.

Our MR-Loop and DR-Loop names emphasize this difference in recurrence
scheduling. They refer to the complete model--training configurations,
not to a controlled recurrence-only ablation; the models additionally
differ in input reinjection, normalization, masking, and readout details
(Appendix~\ref{app:architectures}). All checkpoints studied in the main experiments use three shared
Transformer blocks, two attention heads per block, width $256$, and no
positional embeddings.

\subsection{Sequential Composition Tasks}
\label{sec:tasks}

All three task families instantiate the same sequential-composition
problem. An example contains an initial state $s_0$ followed by $H$
operator tokens,
\[
    [s_0,o_1,\ldots,o_H],
    \qquad
    s_k = T_{o_k}(s_{k-1}), \quad k=1,\ldots,H,
\]
which induce the trajectory
\[
    s_0 \xrightarrow{o_1} s_1
        \xrightarrow{o_2} \cdots
        \xrightarrow{o_H} s_H .
\]
The model observes only the initial state and operators, and is trained
to predict the final state $s_H$. Intermediate states
$s_1,\ldots,s_{H-1}$ are available only for analysis and receive no
training supervision. We call $H$ the \emph{task horizon}; under our
experimental convention, the number of real input tokens is
$n=H+1$.

\begin{table}[ht]
\caption{\textbf{Sequential composition tasks.}
All tasks repeatedly apply operators to a running state but differ in
their state spaces and transition families.}
\centering
\small
\setlength{\tabcolsep}{5pt}
\begin{tabular}{@{}llll@{}}
\toprule
\textbf{Task}
& \textbf{State}
& \textbf{Operator / transition}
& \textbf{Varied parameters} \\
\midrule

Polynomial
& $s_k\in\mathbb{Z}_M$
& $x_k:\;
   s_k=(s_{k-1}x_k+1)\bmod M$
& $M$ \\

FSC
& $s_k\in[S]$
& $o_k:\;
   s_k=f_{o_k}(s_{k-1})$
& $S,F,\alpha$ \\

KG
& $e_k\in[E]$
& $r_k:\;
   e_k=r_k(e_{k-1})$
& $E,R,D$ \\

\bottomrule
\end{tabular}
\label{tab:tasks}
\end{table}

For \textbf{Polynomial}, $x_k$ is an integer multiplier modulo $M$.
For \textbf{finite-state composition (FSC)}, each $o_k$ indexes one of
$F$ functions over $S$ states; a fraction $\alpha$ are random
permutations and the remainder are non-bijective random maps.
For \textbf{knowledge-graph traversal (KG)}, states are entities and
operators are relations; relation maps are random permutations over
$E$ entities, with $D$ valid outgoing relations retained per entity. See \Cref{tab:tasks} for details. Thus, increasing $H$ requires additional applications of the same
task-level transition rule, while the complete ground-truth state
trajectory remains available for mechanistic analysis.

\section{RQ1: How do looped transformers achieve length generalization?}
\label{sec:RQ1}

To characterize the mechanisms underlying length generalization in MR-Loop and DR-Loop, we focus our mechanistic analysis on the knowledge-graph task, using 3-layer models with 2 attention heads per layer and residual dimension \(d_{\mathrm{model}}=256\). Models are trained on inputs up to length \(n=12\) and evaluated on longer sequences to study extrapolation beyond the training horizon. 

\subsection{MR-Loop learns a shared recurrent circuit}
\label{sec:mr_loop_q1}

\begin{figure}[t]
    \centering
    \includegraphics[width=1\linewidth]{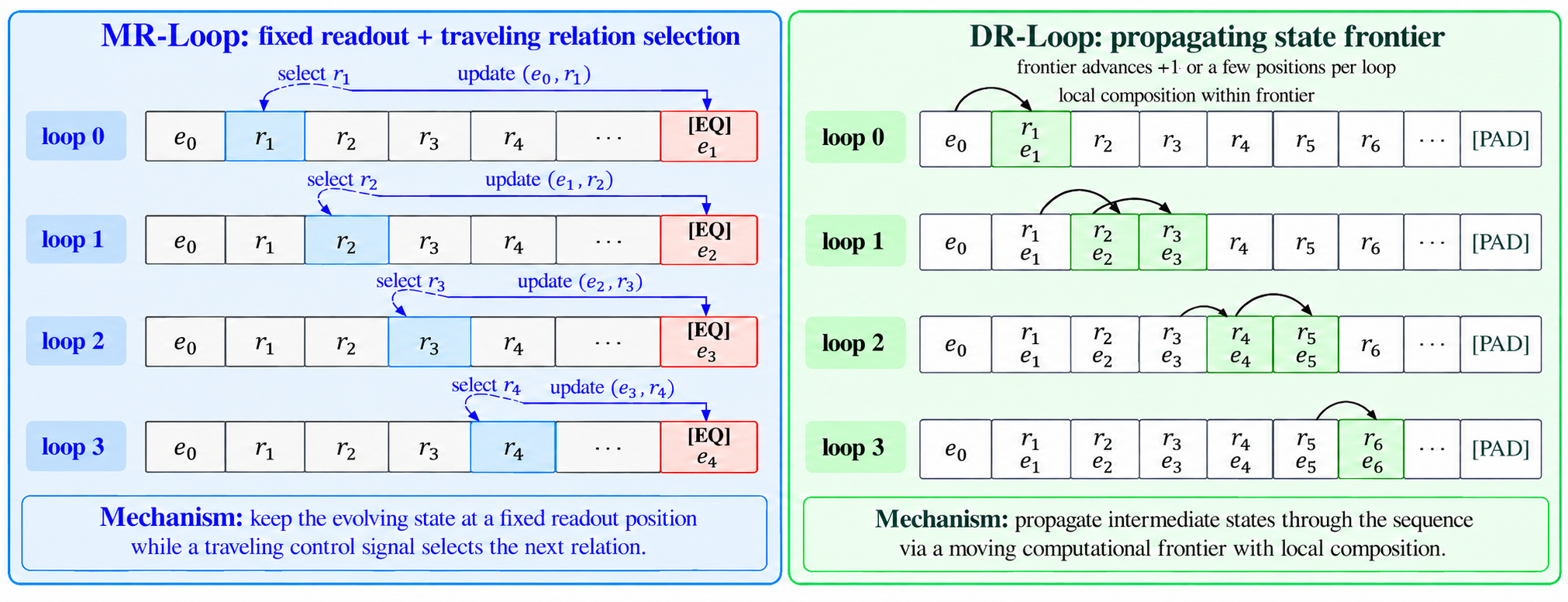}
\caption{
\textbf{Distinct length-generalization mechanisms in looped Transformers}. MR-Loop uses a fixed readout with advancing relation selection; DR-Loop advances a state-propagation frontier.
}
    \label{fig:mechanism_fig}
\end{figure}

Mechanistic analysis reveals that MR-Loop stores the intermediate entity at a fixed readout position, while adjacent-token interactions and a transferable progress cue advance relation selection across recurrent steps. The illustrated mechanism is in \Cref{fig:mechanism_fig}. 

\textbf{Attention analysis.} We use attention weights to track how information is routed across token positions as recurrence proceeds; formal conventions are given in \Cref{app:attention_conventions}. We identify two recurring attention patterns in the MR-loop transformer. The first is a \emph{traveling previous-token} pattern: at loop $t$, the query at $r_{t+2}$ attends strongly to $r_{t+1}$, with this localized attention peak shifting along the sequence across successive loops (\Cref{fig:mrloop_l3h2_kg_l0h1} in \Cref{app:mrloop-kg-attention}). The second is a \emph{fixed-query tracking} pattern: the query remains at the \texttt{[EQ]} position (the answer readout position) while its attention shifts to $r_{t+1}$ at loop $t$ (\Cref{fig:mrloop_l3h2_kg_l2h1} in \Cref{app:mrloop-kg-attention}). In both cases, the attended position advances by one token per loop. This progression persists beyond the maximum training length of $12$ and continues to the terminal \texttt{[EQ]} position. These observations suggest that the model learns an iterative attention mechanism that extends beyond the positions encountered during training, providing a plausible mechanism for length generalization. 

\textbf{Intermediate value tracking at [EQ].} 
We use the logit lens \citep{nostalgebraist2020interpreting} to test which task-relevant state is directly decodable from each residual representation under the model's own output readout (see \Cref{app:logit-lens} for conventions). The attention patterns above suggest an iterative computation. We next investigate whether intermediate values are recoverable during this computation and where they are represented. At each loop $t$, we measure whether the highest-probability prediction at positions $r_t$, $r_{t+1}$, and \texttt{[EQ]} matches the ground-truth intermediate entity $e_{t+1}$. As shown in \Cref{fig:mrloop-kg-intermediate-track} from \Cref{app:mrloop-kg-intermediate}, predictions at \texttt{[EQ]} recover $e_{t+1}$ with perfect accuracy, whereas accuracy at $r_t$ and $r_{t+1}$ remains low. This finding suggests that \texttt{[EQ]} maintains the evolving intermediate state as computation progresses across loops. Moreover, recovery accuracy at \texttt{[EQ]} remains at $100\%$ beyond the maximum training length of $12$ when evaluated on length-$15$ inputs. 

\textbf{Causal tests of iterative composition.}
Together, the attention and decoding results suggest a specific compositional mechanism: at loop $t$, \texttt{[EQ]} carries the current entity $e_t$, while the recurrent attention pattern selects the next relation $r_{t+1}$, producing
$e_{t+1}=r_{t+1}(e_t)$ (see illustration in \Cref{fig:mechanism_fig}). We test this hypothesis causally using complementary state-swap interventions, which replace selected internal states with \textit{donor activations} and test whether downstream computation follows the corresponding counterfactual (see \Cref{app:patching-conventions} for details). First, changing the residual states of relation tokens redirects attention and produces the counterfactual entity obtained by applying the selected relation to the incoming entity at \texttt{[EQ]}. Second, replacing the \texttt{[EQ]} residual with another example's state produces the recipient's next relation applied to the donor's entity. Both effects occur with high accuracy at hop indices beyond the maximum training length of $12$. These interventions provide causal support for the proposed mechanism: relation selection determines the operation, while \texttt{[EQ]} supplies the entity on which it acts (see \Cref{app:mrloop-kg-state-swaps} for full results).

\paragraph{A transferable state cue supports length generalization.} The state-swap experiments in \Cref{app:mrloop-kg-state-swaps} show that relation-token states causally influence which relation is selected. We identify a shared component of these states that distinguishes a relation that is ready to be selected from one that has already been consumed.
For each residual site $s$, we estimate a
\textbf{\emph{live--consumed} direction $v_s$} from the average
difference between a relation token's \textbf{``live''} state at its 
scheduled fetch loop $k-1$ and its \textbf{``consumed''}  state two loops
later, at $k+1$. 
This direction causally controls selection in the tested
settings. Replacing the scheduled key's live residual
$h_s^{\mathrm{live}}$ with its later residual
$h_s^{\mathrm{consumed}}$ causes selection to skip that key.
Using $h_s^{\mathrm{consumed}}+v_s$ restores selection of
the same key and the corresponding entity transition;
using $h_s^{\mathrm{live}}-v_s$ disrupts both.
The input relation tokens remain unchanged throughout.

The same directions, estimated from positions
$k=5,\ldots,11$, also work at later positions on the same
walks. At $k=14$ on length-$20$ inputs, the consumed-state
replacement reduces correct-hop accuracy to $9.4\%$.
Adding $v_s$ restores accuracy to $93.8\%$, matching native
performance, while the tested random direction of the
same norm leaves accuracy at $9.4\%$.
Thus, a shared state cue can make a relation key selectable
beyond the training horizon.
Complementary attention ablations and entity swaps support
a mechanism in which interactions between adjacent relation
tokens advance selection, while \texttt{[EQ]} maintains and
updates the intermediate entity via $e_{t+1}=r_{t+1}(e_t)$. Together, these results support length generalization
through reuse of a relation-selection cue and an
entity-transition computation at additional positions. More details on the live-consumed direction $v_s$ and its corresponding experiments can be found in \Cref{app:mrloop-kg-length-mechanism}.

\subsection{DR-Loop learns a propagating state frontier}
\label{sec:dr-mechanism}


\textbf{DR-Loop propagates intermediate states through an advancing computational frontier.}
In contrast to MR-Loop, which maintains the evolving intermediate state at a fixed readout position, DR-Loop distributes computation across the sequence. Our analysis reveals a \emph{propagating state frontier}: as recurrence proceeds, intermediate entity $e_k$ becomes represented at relation position $k$, and this frontier advances toward later positions across recurrent iterations. Near the frontier, the model combines an upstream entity representation $e_{k-1}$ with the local relation $r_k$ to produce the next state, $e_k = f_{r_k}(e_{k-1})$.


\textbf{Attention propagation beyond the training length.}
As illustrated by the attention patterns in two representative heads in Figures~\ref{fig:drloop-self-attn} and~\ref{fig:drloop-prev-attn} from \Cref{app:drloop_detailed_mechanism_attention}, a localized region of elevated attention progresses toward later sequence positions along the main diagonal and the first subdiagonal, corresponding to attention to the current and immediately preceding positions, respectively. This region spans approximately two to three neighboring positions and advances by roughly one to two positions per recurrent iteration during the propagation phase. After reaching the final relation positions, the localized pattern weakens in subsequent iterations. These observations are consistent with an iterative composition mechanism in which a relation position integrates locally represented information with an intermediate entity representation from an upstream position, enabling computation to propagate along the relation sequence. Moreover, these attention patterns continue to progress on length-15 inputs, beyond the maximum training length of 12, suggesting that the learned recurrent attention dynamics extrapolate beyond the training range and support length generalization. 

\textbf{Propagation of intermediate states beyond the training depth.} Applying the logit lens to intermediate residual states shows that entity $e_k$ becomes decodable at the corresponding relation position $k$. On 14-hop (length-15) queries, entities within the training range ($1 \leq k \leq 11$) reach peak probabilities of approximately $0.95$--$1.00$, while $e_{12}$ and $e_{13}$ remain strongly decodable at approximately $0.90$--$0.92$. We report the maximum over recurrent iterations at each position, since decodability can decrease after its peak. As shown in \Cref{fig:drloop_l3h2_kg_logitlens}, this advancing frontier of entity decodability tracks the localized attention progression and extends beyond the maximum training length, providing representational evidence that recurrent computation is reused at OOD length.

\textbf{Recurrent state propagation and causal evidence.}
The attention and logit-lens analyses motivate a recurrent
state-propagation mechanism: near the computational frontier, the model
combines information about the upstream entity $e_{k-1}$ with
relation-dependent information at position $k$ to produce a representation of $e_k=f_{r_k}(e_{k-1})$ at that position. Repeated application of this computation could account for the observed progression toward later
relation positions. We test this hypothesis using state-swap activation
patching on 64 queries requiring 14 hops, beyond the maximum training
depth of 11 hops. Donors have a different starting entity but identical
relations, defining a counterfactual continuation
$e'_j=f_{r_j}(e'_{j-1})$. Patching position $5$ makes $e'_6$ and $e'_{14}$ the final-loop
decoded argmax in $90.6\%$ and $71.9\%$ of examples, respectively;
the final PAD readout predicts the counterfactual answer in $73.4\%$.
Direct intervention beyond the training depth also changes the
computation: patching position $12$ makes $e'_{13}$ the decoded argmax
in $56.25\%$ of examples.
These results show that intermediate residual states causally influence
subsequent composition and final predictions, supporting recurrent
state propagation beyond the training depth.
Appendix~\ref{app:drloop-state-swap} provides intervention details,
controls, and full results. 

\paragraph{Length generalization through reuse of a local composition rule.} 
We test whether the model reuses entity--relation composition at
unseen hops by transplanting the same fresh representation of an
early entity, $e_2$, to several later frontier positions $F$.
Under this hypothesis, position $F+1$ should represent the
counterfactual $f_{r_{F+1}}(e_2)$, obtained by applying the relation
at the receiving position to the transplanted entity. We assess
this prediction using the logit lens. For example, after
transplanting $e_2$ to position $12$, the counterfactual
$f_{r_{13}}(e_2)$ reaches a peak batch-mean probability of $0.336$
at position $13$, compared with $0.037$ for the donor's original
successor, $f_{r_3}(e_2)$. This preference for the receiving
position's relation occurs beyond the maximum training depth
of $11$ hops. Similar counterfactual responses at other tested
frontiers (\Cref{app:dr-loop-kg-phase-reuse}) further support reuse of entity--relation
composition beyond the training range. 


Crucially, successful reuse depends not only on \textbf{\emph{what}} entity is represented, but also on the
\textbf{\emph{computational state}} of that representation. A newly computed (\textbf{\emph{fresh}}) entity state
supports subsequent composition, whereas the same state after additional recurrent processing
(\textbf{\emph{aged}}) can remain strongly decodable while becoming nearly unusable for the next relation
application. For example, an aged representation of $e_2$ remains decodable with $96.9\%$
argmax accuracy, yet its next-hop score falls from $0.650$ when fresh to $0.038$ when aged.
We identify a shared \textbf{fresh--aged residual direction} that causally affects its usefulness for composition:
adding the direction to an aged state restores its next-hop score to $0.698$, whereas subtracting
it from a fresh state reduces the score to $0.031$. Importantly, the same direction transfers across
source positions and remains effective at unseen hop indices. We refer to this transferable
residual-state property as \emph{computational phase}. Together, these results suggest that
DR-Loop's length generalization relies on reusing a shared local composition rule while maintaining
intermediate states in a representation that remains computationally compatible with that rule
(more details in \Cref{app:dr-loop-kg-phase-reuse}).

\subsection{A Common Principle: semantic content is not sufficient for recurrent reuse}
\label{sec:computation_status}

\begin{figure}[t]
    \centering
    \includegraphics[width=1\linewidth]{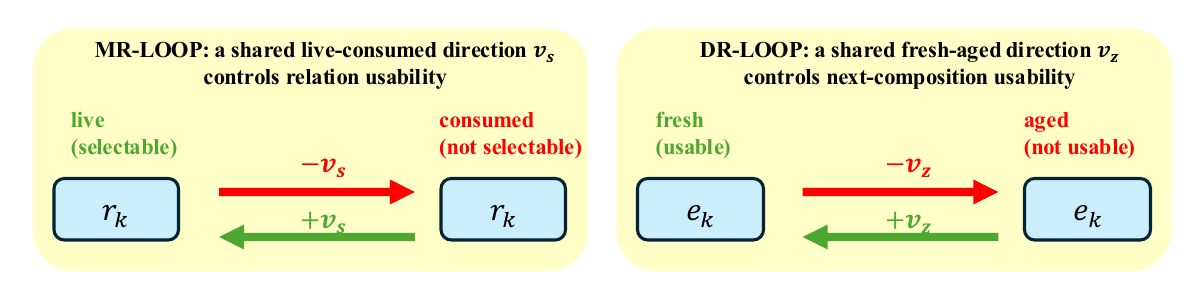}
\caption{
\textbf{Computational status controls recurrent reuse.} Live–consumed relation states govern selection in MR-Loop, while fresh–aged intermediate states govern composition in DR-Loop. 
}
    \label{fig:computation_status}
\end{figure}

Although MR-Loop and DR-Loop implement different recurrent algorithms, both reveal
a common representational principle: successful reuse depends not only on
\emph{what} information a state represents, but also on whether that state remains
\emph{computationally usable}. We refer to this latter property broadly as
\emph{computational status}. In MR-Loop, computational status is expressed through the live--consumed state of
relation representations. Holding relation identity fixed, replacing a live
relation residual with its consumed state prevents the corresponding transition,
whereas adding a shared live--consumed direction restores relation selection and
the correct update, including at positions beyond the training horizon. In
DR-Loop, the distinction appears in intermediate entities: an aged state can remain
strongly decodable as the correct entity while becoming nearly unusable for the
next relation application. A shared fresh--aged direction causally restores this
ability and transfers across positions and unseen hop indices. \Cref{fig:computation_status} illustrates this shared computational-status
principle: task-relevant content can remain represented while its usability for subsequent recurrent computation changes. 

These results separate \emph{semantic representation} from
\emph{computational usability}. Preserving the correct intermediate value is
therefore not sufficient for recurrent reasoning; the value must remain in a
representation compatible with the next application of the learned transition.
MR-Loop and DR-Loop realize this requirement differently---through live relation
selection and fresh intermediate-state propagation, respectively---but both
suggest that maintaining computational status is a key condition for recurrent
length generalization.

\subsection{Generality across tasks and configurations}

The identified mechanisms are not restricted to the primary KG
checkpoints. Separately trained MR-Loop models on KG, Polynomial, and
FSC reproduce the fixed-\texttt{[EQ]} accumulator, advancing input
selection, and the causal live--consumed routing effect.  For DR-Loop,
where Section~\ref{sec:RQ3} shows that other task families do not always
exhibit a faithful task-aligned intermediate trajectory, two additional
KG configurations reproduce the advancing state frontier and the
fresh--aged dependence of subsequent composition.  Thus, the broader
computational-status organization recurs across tasks and independently
trained configurations, while its concrete realization remains
task- and checkpoint-dependent. See
\Cref{app:mechanism-generality} for full results. 


\section{RQ2: Why does length generalization fail at extremely long horizons?} 
\label{sec:RQ2}

Our knowledge-graph experiments identify two coupled causes of
extreme-length failure: \textbf{recurrent state propagation becomes less reliable at extreme lengths, while the learned dynamics provide limited recovery from disrupted computation}. Errors can therefore affect subsequent updates, and additional recurrence need not restore the correct trajectory. The accompanying \Cref{app:extreme_length_failure} gives the experimental protocols and quantitative results.

\textbf{Degradation of the states that support computation.} In MR-Loop, longer prompts impair the recurrent \texttt{[EQ]} state even at shared hop indices. Transplanting a same-loop \texttt{[EQ]} state from a shorter prompt restores local transitions while leaving the longer prompt's relation keys unchanged. For example, replacing the incoming \texttt{[EQ]} residual of a length-30 run with the same-loop residual from its matched length-20 prefix, before block 0, raises mean native next-hop accuracy over keys $12\leq k\leq19$ from $5.5\%$ to $83.6\%$ (\Cref{app:mr_loop_RQ2_eq}). 
The live–consumed relation cue also weakens at greater depth: its mean residual difference projected onto a fixed unit direction estimated at trained positions remains near $28$ through key $15$, then declines to $16.1$ at key $21$ and $9.1$ at key $23$ (\Cref{app:mr_loop_RQ2_procue}). In DR-Loop, intermediate-state confidence and subsequent composition deteriorate along the chain. For example, on 29-hop inputs, DR-Loop’s intermediate confidence declines with hop depth—mean per-example peak correct-entity probability within the evaluation windows falls from \(0.982\) at hop 5 to \(0.713\) at hop 16 (\Cref{app:dr_loop_RQ2}). In DR-Loop, confidently decodable intermediates are associated with stronger next-hop predictions (\Cref{app:dr_loop_RQ2}), but continuation confidence still declines with depth, consistent with intermediate-state degradation contributing to generalization failure at extreme lengths. 



\textbf{State organization and computational status constrain self-correction.}
\Cref{sec:RQ1} identifies distinct constraints on revisiting failed computations. MR-Loop uses \texttt{[EQ]} as a single evolving entity register, successively overwriting its active intermediate as relation selection advances. The identified mechanism provides no demonstrated means of retrieving a preserved, valid earlier entity state. Following corruption, subsequent updates operate on the altered register; retrying a failed hop would require reconstructing a valid predecessor and realigning the live relation. DR-Loop retains intermediates across token positions, but this distributed trace does not automatically provide usable checkpoints. As \Cref{sec:computation_status} shows, an entity can remain decodable at position \(k\) after becoming aged and ineffective for composition. Recomputing an erroneous intermediate may therefore require restoring its predecessor's freshness, preparing the receiving position, and redirecting computation to that hop before propagating the correction downstream (more discussion in \Cref{app:computational_status_recovery}). These state organizations help explain why additional recurrence need not implement repair: MR-Loop replaces earlier active states, while DR-Loop can retain earlier values without retaining their usability for replay. These results indicate that self-correction is limited. 

\textbf{Disruption can persist through subsequent updates.} In DR-Loop, advancing an upstream residual by one recurrent pass causes persistent damage to later intermediates after the intervention ends (\Cref{app:dr_loop_RQ2}). In MR-Loop, a separate analysis of native first-error trajectories finds no sustained recovery in the evaluated cohorts. Isolated correct later readouts occur, but do not establish restoration of the remaining computation (\Cref{app:mr_firsthopfailure_recovery}). 

\textbf{Consequences for length generalization.} Let \(C_k\) denote correctness of the intermediate readout at hop \(k\) under a specified evaluation protocol. Define the conditional first-error hazard \(h_k(n)=\Pr(\neg C_k\mid C_1,\ldots,C_{k-1},n)\). The probability of a completely correct measured trajectory is:

\[
S_K(n)=\Pr\!\left(\bigcap_{k=1}^{K}C_k\mid n\right)
=\prod_{k=1}^{K}\left(1-h_k(n)\right).
\]

This identity requires no independence assumption. When intact trajectories usually finish correctly and recovery after errors is rare, final-answer accuracy approximately follows this survival probability. Even a constant conditional error \(h>0\) yields geometric decay, \(S_K=(1-h)^K\). If the conditional error probability increases with depth or input length, trajectory survival can decline more steeply than under a fixed-hazard model. This provides a qualitative account of how deteriorating computational states and limited recovery can amplify length-generalization failure. Thus, in the studied models, extreme-length generalization fails when repeated updates degrade the states needed for subsequent computation and the learned recurrence does not reliably repair them: longer trajectories create more opportunities for local errors whose effects persist through the remaining computation. Note that this survival analysis applies most directly to tasks without shortcut solutions, where success depends on preserving the intermediate trajectory.  


\section{RQ3: Does successful length generalization imply faithful iterative computation?}
\label{sec:RQ3}

We test whether successful length generalization is accompanied by recovery of the ground-truth intermediate trajectory, allowing recurrent passes and task transitions to be misaligned. 

\textbf{Length-generalization success depends on task structure and model configuration.}
We first evaluate whether length generalization emerges across task
families and problem configurations, using a 3-layer, 2-head model as
our default architecture. On polynomial iteration, we vary the modulus
over $ m \in \{11,19,25,29,30,35,40,50\}. $
MR-Loop length-generalizes for all tested composite moduli and for
$m=11$, but not for the other tested
prime moduli. DR-Loop also fails on the tested prime-modulus settings
and succeeds on several composite moduli, although not uniformly.
On FSC, MR-Loop length-generalizes across
all tested configurations, while DR-Loop becomes substantially less
robust when the number of transition functions is increased. On the
KG task, whose relation maps are permutations in our
construction, MR-Loop succeeds across all six tested configurations,
whereas DR-Loop succeeds in three. Full results are  in
Table~\ref{tab:generalization_summary} from \Cref{app:RQ3_additional_length_tasks}.

\textbf{Successful extrapolation does not guarantee recovery of the intended intermediate trajectory.} For MR-Loop on knowledge-graph traversal, the evolving entity is recovered
nearly perfectly at the recurrent readout, consistent with the
mechanism identified in \Cref{sec:RQ1}. Similarly, fully bijective FSC
and the successful $m=11$ polynomial setting expose substantially more
complete task-aligned trajectories. In contrast, on polynomial
iteration with $m=50$ and more difficult partially compressive FSC
settings, the model can length-generalize even though earlier
ground-truth intermediate states are only weakly recovered and the
trajectory becomes clear mainly near the final updates
(Figure~\ref{fig:mrloop_intermediates_acc}). DR-Loop exhibits an even clearer separation between behavioral success
and intermediate-state recovery. In composite-modulus polynomial tasks
and easier FSC configurations, the model can produce correct
final answers while many ground-truth intermediates remain weak
or unrecoverable along the recurrent trajectory
(Figure~\ref{fig:drloop_intermediates_weight_fsc}). 

\textbf{Task compressibility modulates both learnability and
intermediate-state faithfulness.}
We interpret these results using the task-level compressibility framework
of \citet{liang2026latent}. Let $G_t$ denote the composition of the
remaining transitions after intermediate state $s_t$. If
$G_t(u)=G_t(v)$ for two possible states $u\neq v$, then distinguishing
$u$ from $v$ is unnecessary for predicting the final answer: a
representation of their suffix-induced equivalence class can suffice.
When $G_t$ is injective, by contrast, distinct running states remain
relevant to the final answer and therefore cannot be irreversibly
merged. Composite-modulus polynomial tasks and FSC with non-bijective
transitions can admit such compression, whereas prime-modulus
polynomial tasks, fully bijective FSC, and our permutation-based KG
construction preserve state distinctions
(Appendix~\ref{app:compressibility}).

This distinction helps organize two empirical patterns. First,
compressive settings often exhibit more robust behavioral length
generalization. The contrast is clearest for polynomial iteration:
both architectures succeed substantially more often on the tested
composite moduli than on the prime moduli. We interpret this not as a
guarantee that compression makes optimization easy, but as a reduction
in the information-preservation burden: compressive tasks admit
coarser sufficient representations and hence additional shortcut
solutions, whereas information-preserving tasks require successful
computation to retain all distinctions relevant to the remaining
composition. Second, the same distinction helps explain the faithfulness gap. 
When compression is available, successful prediction need not recover
every earlier ground-truth intermediate, consistent with the weak
early-state recovery observed in several composite-polynomial and
partially compressive FSC settings. In information-preserving settings,
this particular lossy strategy is unavailable; successful MR-Loop
models on KG, fully bijective FSC, and polynomial $m=11$ accordingly
expose substantially more complete task-aligned trajectories.

\textbf{Takeaway.}
Taken together, these results distinguish \emph{behavioral length
generalization} from \emph{intermediate-state faithfulness}. Both
MR-Loop and DR-Loop can generalize beyond the training horizon without
consistently exposing the complete ground-truth intermediate trajectory
under our readout analysis. Task-side compressibility constrains the
available solution space by determining whether lossy sufficient
representations are possible, but it alone does not determine behavioral
success or the learned algorithm. Successful length generalization
therefore demonstrates behavioral transfer to longer horizons, but does
not, by itself, establish that the model executes the intended
state-by-state procedure.

\section{Conclusion}


We provide a mechanistic study of how looped Transformers reuse computation
beyond the training horizon. MR-Loop and DR-Loop achieve length
generalization through distinct recurrent algorithms---a fixed accumulator
with advancing operator selection versus a propagating state frontier---yet
reveal a common requirement: representing the correct information is not
enough; that information must remain in a computationally usable state for
the next transition. At extreme lengths, these computation-ready states
degrade and the learned dynamics provide limited correction, allowing local
errors to persist and accumulate. We further show that successful length
generalization need not imply faithful recovery of the intended intermediate
trajectory, particularly when task structure permits lossy shortcuts.
Together, these results suggest that reliable recurrent reasoning depends not
simply on allocating more test-time computation, but on preserving and
refreshing the internal states that make that computation reusable. Future
work should do looped Transformer structure ablation and test these principles at larger scales and in more naturalistic
reasoning settings.




\bibliographystyle{iclr2027_conference}
\bibliography{iclr2027_conference}

@inproceedings{fan2024looped,
  author = {Fan, Ying and Du, Yilun and Ramchandran, Kannan and Lee, Kangwook},
  title = {Looped Transformers for Length Generalization},
  booktitle = {The Thirteenth International Conference on Learning Representations},
  year = {2025},
  url = {https://arxiv.org/abs/2409.15647}
}

@inproceedings{kohli2026loop,
  author = {Kohli, Harsh and Parthasarathy, Srinivasan and Sun, Huan and Yao, Yuekun},
  title = {Loop, Think, \& Generalize: Implicit Reasoning in Recurrent-Depth Transformers},
  booktitle = {Conference on Language Modeling},
  year = {2026},
  url = {https://arxiv.org/abs/2604.07822}
}

@inproceedings{dehghani2019universal,
  author = {Dehghani, Mostafa and Gouws, Stephan and Vinyals, Oriol and Uszkoreit, Jakob and Kaiser, Lukasz},
  title = {Universal Transformers},
  booktitle = {The Seventh International Conference on Learning Representations},
  year = {2019},
  url = {https://openreview.net/forum?id=HyzdRiR9Y7}
}

@inproceedings{giannou2023looped,
  author = {Giannou, Angeliki and Rajput, Shashank and Sohn, Jy-Yong and Lee, Kangwook and Lee, Jason D. and Papailiopoulos, Dimitris},
  title = {Looped Transformers as Programmable Computers},
  booktitle = {Proceedings of the 40th International Conference on Machine Learning},
  pages = {11398--11442},
  year = {2023},
  volume = {202},
  series = {Proceedings of Machine Learning Research},
  publisher = {PMLR},
  url = {https://proceedings.mlr.press/v202/giannou23a.html}
}

@inproceedings{yang2024looped,
  author = {Yang, Liu and Lee, Kangwook and Nowak, Robert D. and Papailiopoulos, Dimitris},
  title = {Looped Transformers Are Better at Learning Learning Algorithms},
  booktitle = {The Twelfth International Conference on Learning Representations},
  year = {2024},
  url = {https://openreview.net/forum?id=HHbRxoDTxE}
}

@inproceedings{xu2025expressive,
  author = {Xu, Kevin and Sato, Issei},
  title = {On Expressive Power of Looped Transformers: Theoretical Analysis and Enhancement via Timestep Encoding},
  booktitle = {Proceedings of the 42nd International Conference on Machine Learning},
  pages = {69613--69646},
  year = {2025},
  volume = {267},
  series = {Proceedings of Machine Learning Research},
  publisher = {PMLR},
  url = {https://proceedings.mlr.press/v267/xu25x.html}
}

@inproceedings{zhou2024algorithms,
  author = {Zhou, Hattie and Bradley, Arwen and Littwin, Etai and Razin, Noam and Saremi, Omid and Susskind, Joshua M. and Bengio, Samy and Nakkiran, Preetum},
  title = {What Algorithms Can Transformers Learn? A Study in Length Generalization},
  booktitle = {The Twelfth International Conference on Learning Representations},
  year = {2024},
  url = {https://openreview.net/forum?id=AssIuHnmHX}
}

@article{zhou2024robust,
  author = {Zhou, Yongchao and Alon, Uri and Chen, Xinyun and Wang, Xuezhi and Agarwal, Rishabh and Zhou, Denny},
  title = {Transformers Can Achieve Length Generalization but Not Robustly},
  journal = {arXiv preprint arXiv:2402.09371},
  year = {2024},
  doi = {10.48550/arXiv.2402.09371},
  url = {https://arxiv.org/abs/2402.09371}
}

@inproceedings{cho2024position,
  author = {Cho, Hanseul and Cha, Jaeyoung and Awasthi, Pranjal and Bhojanapalli, Srinadh and Gupta, Anupam and Yun, Chulhee},
  title = {Position Coupling: Improving Length Generalization of Arithmetic Transformers Using Task Structure},
  booktitle = {Advances in Neural Information Processing Systems},
  volume = {37},
  pages = {22233--22315},
  year = {2024},
  publisher = {Curran Associates, Inc.},
  doi = {10.52202/079017-0701},
  url = {https://proceedings.neurips.cc/paper_files/paper/2024/hash/27aa3a0e6d63db269977bb2df5607cb8-Abstract-Conference.html}
}

@article{sabbaghi2024structural,
  author = {Sabbaghi, Mahdi and Pappas, George J. and Hassani, Hamed and Goel, Surbhi},
  title = {Explicitly Encoding Structural Symmetry Is Key to Length Generalization in Arithmetic Tasks},
  journal = {arXiv preprint arXiv:2406.01895},
  year = {2024},
  doi = {10.48550/arXiv.2406.01895},
  url = {https://arxiv.org/abs/2406.01895}
}

@article{blayney2026mechanistic,
  author = {Blayney, Hugh and Arroyo, {\'A}lvaro and Obando-Ceron, Johan and Castro, Pablo Samuel and Courville, Aaron and Bronstein, Michael M. and Dong, Xiaowen},
  title = {A Mechanistic Analysis of Looped Reasoning Language Models},
  journal = {arXiv preprint arXiv:2604.11791},
  year = {2026},
  doi = {10.48550/arXiv.2604.11791},
  url = {https://arxiv.org/abs/2604.11791}
}

@article{labovich2026stability,
  author = {Labovich, Asher},
  title = {Stability and Generalization in Looped Transformers},
  journal = {arXiv preprint arXiv:2604.15259},
  year = {2026},
  doi = {10.48550/arXiv.2604.15259},
  url = {https://arxiv.org/abs/2604.15259}
}

@misc{liang2026latent,
title={Do Latent-CoT Models Think Step-by-Step? A Mechanistic Study on Sequential Reasoning Tasks},
author={Jia Liang and Liangming Pan},
year={2026},
eprint={2602.00449},
archivePrefix={arXiv},
primaryClass={cs.AI},
url={https://arxiv.org/abs/2602.00449},
}

@article{nostalgebraist2020interpreting,
  title={Interpreting GPT: The logit lens},
  author={nostalgebraist, Rob},
  journal={Blog Post},
  year={2020}
}

@inproceedings{meng2022locating,
author = {Meng, Kevin and Bau, David and Andonian, Alex and Belinkov, Yonatan},
booktitle = {Advances in Neural Information Processing Systems},
doi = {10.52202/068431-1262},
editor = {S. Koyejo and S. Mohamed and A. Agarwal and D. Belgrave and K. Cho and A.
Oh},
pages = {17359--17372},
publisher = {Curran Associates, Inc.},
title = {Locating and Editing Factual Associations in GPT},
url =
{https://proceedings.neurips.cc/paper_files/paper/2022/file/6f1d43d5a82a37e89b0665b33
bf3a182-Paper-Conference.pdf},
volume = {35},
year = {2022}
}

@article{wang2022interpretability,
  title={Interpretability in the wild: a circuit for indirect object identification in GPT-2 small},
  author={Wang, Kevin and Variengien, Alexandre and Conmy, Arthur and Shlegeris, Buck and Steinhardt, Jacob},
  journal={arXiv preprint arXiv:2211.00593},
  year={2022}
}

@inproceedings{schwarzschild2021algorithm,
  author = {Schwarzschild, Avi and Borgnia, Eitan and Gupta, Arjun and Huang, Furong and Vishkin, Uzi and Goldblum, Micah and Goldstein, Tom},
  title = {Can You Learn an Algorithm? Generalizing from Easy to Hard Problems with Recurrent Networks},
  booktitle = {Advances in Neural Information Processing Systems},
  volume = {34},
  pages = {6695--6706},
  year = {2021},
  publisher = {Curran Associates, Inc.},
  url = {https://proceedings.neurips.cc/paper/2021/hash/3501672ebc68a5524629080e3ef60aef-Abstract.html}
}

@inproceedings{thomm2024limits,
  author = {Thomm, Jonathan and Camposampiero, Giacomo and Terzic, Aleksandar and Hersche, Michael and Sch{\"o}lkopf, Bernhard and Rahimi, Abbas},
  title = {Limits of Transformer Language Models on Learning to Compose Algorithms},
  booktitle = {Advances in Neural Information Processing Systems},
  volume = {37},
  year = {2024},
  publisher = {Curran Associates, Inc.},
  doi = {10.52202/079017-0245},
  url = {https://proceedings.neurips.cc/paper_files/paper/2024/hash/0e797d5139ad94fc2dc2080c09119f29-Abstract-Conference.html}
}

@inproceedings{anil2022exploring,
  author = {Anil, Cem and Wu, Yuhuai and Andreassen, Anders and Lewkowycz, Aitor and Misra, Vedant and Ramasesh, Vinay and Slone, Ambrose and Gur-Ari, Guy and Dyer, Ethan and Neyshabur, Behnam},
  title = {Exploring Length Generalization in Large Language Models},
  booktitle = {Advances in Neural Information Processing Systems},
  volume = {35},
  pages = {38546--38556},
  year = {2022},
  publisher = {Curran Associates, Inc.},
  doi = {10.52202/068431-2793},
  url = {https://proceedings.neurips.cc/paper_files/paper/2022/hash/fb7451e43f9c1c35b774bcfad7a5714b-Abstract-Conference.html}
}

@inproceedings{huang2025formal,
  author = {Huang, Xinting and Yang, Andy and Bhattamishra, Satwik and Sarrof, Yash and Krebs, Andreas and Zhou, Hattie and Nakkiran, Preetum and Hahn, Michael},
  title = {A Formal Framework for Understanding Length Generalization in Transformers},
  booktitle = {The Thirteenth International Conference on Learning Representations},
  year = {2025},
  url = {https://openreview.net/forum?id=U49N5V51rU}
}

@article{hahn2020theoretical,
  author = {Hahn, Michael},
  title = {Theoretical Limitations of Self-Attention in Neural Sequence Models},
  journal = {Transactions of the Association for Computational Linguistics},
  volume = {8},
  pages = {156--171},
  year = {2020},
  publisher = {MIT Press},
  doi = {10.1162/tacl_a_00306},
  url = {https://aclanthology.org/2020.tacl-1.11/}
}

@article{geiping2025scaling,
  author = {Geiping, Jonas and McLeish, Sean and Jain, Neel and Kirchenbauer, John and Singh, Siddharth and Bartoldson, Brian R. and Kailkhura, Bhavya and Bhatele, Abhinav and Goldstein, Tom},
  title = {Scaling up Test-Time Compute with Latent Reasoning: A Recurrent Depth Approach},
  journal = {arXiv preprint arXiv:2502.05171},
  year = {2025},
  doi = {10.48550/arXiv.2502.05171},
  url = {https://arxiv.org/abs/2502.05171}
}

@article{yang2026stabilizing,
  author = {Yang, Xiao-Wen and Han, Ziyu and Zhang, Xi-Hua and Wei, Wen-Da and Shao, Jie-Jing and Guo, Lan-Zhe and Li, Yu-Feng},
  title = {Stabilizing Recurrent Dynamics for Test-Time Scalable Latent Reasoning in Looped Language Models},
  journal = {arXiv preprint arXiv:2605.26733},
  year = {2026},
  doi = {10.48550/arXiv.2605.26733},
  url = {https://arxiv.org/abs/2605.26733}
}

@inproceedings{saunshi2025latent,
  author = {Saunshi, Nikunj and Dikkala, Nishanth and Li, Zhiyuan and Kumar, Sanjiv and Reddi, Sashank J.},
  title = {Reasoning with Latent Thoughts: On the Power of Looped Transformers},
  booktitle = {The Thirteenth International Conference on Learning Representations},
  year = {2025},
  url = {https://openreview.net/forum?id=din0lGfZFd}
}

@article{fan2026bridging,
  author = {Fan, Ying and Svete, Anej and Lee, Kangwook},
  title = {Bridging the Gap Between Latent and Explicit Reasoning with Looped Transformers},
  journal = {arXiv preprint arXiv:2606.31779},
  year = {2026},
  doi = {10.48550/arXiv.2606.31779},
  url = {https://arxiv.org/abs/2606.31779}
}

@inproceedings{turpin2023unfaithful,
  author = {Turpin, Miles and Michael, Julian and Perez, Ethan and Bowman, Samuel R.},
  title = {Language Models Don't Always Say What They Think: Unfaithful Explanations in Chain-of-Thought Prompting},
  booktitle = {Advances in Neural Information Processing Systems},
  volume = {36},
  pages = {74952--74965},
  year = {2023},
  publisher = {Curran Associates, Inc.},
  doi = {10.52202/075280-3275},
  url = {https://proceedings.neurips.cc/paper_files/paper/2023/hash/ed3fea9033a80fea1376299fa7863f4a-Abstract-Conference.html}
}

@article{lanham2023faithfulness,
  author = {Lanham, Tamera and Chen, Anna and Radhakrishnan, Ansh and Steiner, Benoit and Denison, Carson and Hernandez, Danny and Li, Dustin and Durmus, Esin and Hubinger, Evan and Kernion, Jackson and Luko{\v{s}}i{\=u}t{\.e}, Kamil{\.e} and Nguyen, Karina and Cheng, Newton and Joseph, Nicholas and Schiefer, Nicholas and Rausch, Oliver and Larson, Robin and McCandlish, Sam and Kundu, Sandipan and Kadavath, Saurav and Yang, Shannon and Henighan, Thomas and Maxwell, Timothy and Telleen-Lawton, Timothy and Hume, Tristan and Hatfield-Dodds, Zac and Kaplan, Jared and Brauner, Jan and Bowman, Samuel R. and Perez, Ethan},
  title = {Measuring Faithfulness in Chain-of-Thought Reasoning},
  journal = {arXiv preprint arXiv:2307.13702},
  year = {2023},
  doi = {10.48550/arXiv.2307.13702},
  url = {https://arxiv.org/abs/2307.13702}
}

\appendix

\section{Looped Transformer architectures}
\label[appendix]{app:architectures}
Both models reuse a stack $B$ of $L$ GPT-2 blocks.
Studied checkpoints have no positional embeddings. We write $x$ for the
token sequence, $e=E(x)$ for token embeddings, $n$ for the number of real
input tokens, and $n-1$ for the hop count.
\subsection{MR-Loop (matched recurrence looped transformer)}
\label{app:mr-loop}
MR-Loop follows \citet{fan2024looped}. Recurrence starts from a zero
residual and mixes the previous state with a fresh copy of $e$ on every
pass (input injection):
\begin{equation}
  h^{(0)}=0,\qquad
  h^{(t)}=B\!\left(h^{(t-1)}+e\right),
  \quad t=1,\ldots,T.
  \label{eq:mr-loop}
\end{equation}
The first pass is $B(e)$. Final layer normalization sits inside $B$, so
$h^{(t)}$ is normalized before being added to $e$ on the next pass.

\paragraph{Matched training.}
An example of length $n$ is supervised after the shared block $B$ has been
applied $T=n$ times. A length curriculum raises the maximum $n$ on a fixed
step schedule (default: from $3$ to $12$, increasing every $1000$ steps).
Online batches mix lengths up to the current maximum. There is no
intermediate-state supervision: the loss is applied only at the answer
position \texttt{EQ}. At evaluation, only the residual after $T=n$
applications of $B$ is passed through the output head.
Matched recurrence therefore couples three quantities: input length,
supervision depth, and evaluation depth.

\subsection{DR-Loop (decoupled recurrence)}
\label[appendix]{app:dr-loop}

DR-Loop follows \citet{kohli2026loop}. The input is
embedded once and later passes apply $B$ without reinjection:
\begin{equation}
  h^{(0)}=e,\qquad
  h^{(t)}=B\!\left(h^{(t-1)}\right),
  \quad t=1,\ldots,R.
  \label{eq:dr-loop}
\end{equation}
There is no final layer norm between passes. After the last pass,
$\hat y=W_U\,\mathrm{LN}_f(h^{(R)})$ with tied weights $W_U=E^\top$.
There is no \texttt{EQ} token; the answer is read
at a right-padded PAD position.
\paragraph{Decoupled training.}
The number of real input tokens $n$ and recurrence depth are trained as separate knobs. A hop
curriculum increases the maximum composition depth when in-distribution
accuracy crosses a threshold (default $0.95$), while each batch draws
\begin{equation}
  R\sim\mathrm{clip}\bigl(\mathrm{Poisson}(\lambda),\,R_{\min},\,R_{\max}\bigr)
\end{equation}
with $\lambda=4$ and $[R_{\min},R_{\max}]=[2,8]$ as our default setup. Thus an $n$-token example is not assigned $R=n$. Evaluation
may fix $R$, including $R$ outside $\{2,\ldots,8\}$, which is the intended
inference-time compute axis. 
Decoupled recurrence therefore does not identify loop index with hop
index. Extra passes can be allocated at test time without changing $n$.
\subsection{Summary}
\label{app:loop-comparison}
Table~\ref{tab:loop-architectures} lists main differences between DR-Loop and MR-loop. Our MR-Loop and DR-Loop labels reflect the most salient recurrence distinction between the two model families: matched versus decoupled recurrent depth. The models also differ in input mixing, normalization, delimiter/readout design, PAD masking, and output-head parameterization; consequently, the mechanistic differences we observe should be interpreted as differences between the complete model/training configurations rather than as effects of recurrence scheduling alone. 

\begin{table}[t]
\centering
\caption{MR-Loop versus DR-Loop in the checkpoints studied here.}
\label{tab:loop-architectures}
\small
\begin{tabular}{lll}
\toprule
& MR-Loop & DR-Loop \\
\midrule
Recurrence depth vs.\ length & matched: loss after $T=n$ passes
  & decoupled: $R$ independent of $n$ \\
Train curriculum & increasing on a fixed step schedule
  & increasing only after required accuracy is met \\
Train depth & length-dependent $T$
  & $R\sim\mathrm{Poisson}$, clipped to $\{2,\ldots,8\}$ \\
Eval depth & $T=n$ (default)
  & chosen $R$, possibly $R\neq n$ \\
Recurrent update & $h^{(t)}=B(h^{(t-1)}+e)$
  & $h^{(t)}=B(h^{(t-1)})$ \\
Final LN in the loop & yes, inside $B$
  & no; $\mathrm{LN}_f$ at readout \\
Output head & untied $W_o$ & tied $W_U=E^\top$ \\
Answer site & \texttt{EQ} & last PAD \\
PAD as attention key & yes & masked \\
\bottomrule
\end{tabular}
\end{table}

\section{Mechanistic Analysis Tools}

\subsection{Attention-analysis conventions}
\label{app:attention_conventions}

For recurrent pass $t$, layer $\ell$, and attention head $h$, let
\[
A^{(t,\ell,h)}_{qk}
=
\operatorname{softmax}_{k}
\left(
\frac{
\langle q^{(t,\ell,h)}_q,
        k^{(t,\ell,h)}_k\rangle
}{\sqrt{d_h}}
+ M_{qk}
\right)
\]
denote the causal self-attention weight from query position $q$ to
key position $k$, where $M_{qk}=-\infty$ for $k>q$.
Each row therefore sums to one over the positions visible to the query.

In the attention maps below, rows denote query positions and columns
denote key positions. Because the Transformer blocks are shared across
recurrent passes, the same attention head is applied repeatedly as
the hidden states evolve. We therefore examine
$\{A^{(t,\ell,h)}\}_{t=0}^{R-1}$ for a fixed head to determine whether
its routing pattern remains stationary or advances through the input
as recurrence proceeds.

We use attention patterns as descriptive evidence about information
routing rather than as standalone evidence of causal computation.
Accordingly, the attention analyses below are paired with
intermediate-state decoding and activation interventions that test
whether the attended representations functionally influence subsequent
computation.

\subsection{Logit-Lens and Intermediate-Decoding Conventions}
\label{app:logit-lens}

We use the logit lens \citep{nostalgebraist2020interpreting} to test whether task-relevant intermediate states are
directly decodable from the model's residual stream using its own output
readout. Let $h^{(t)}_{b,j}\in\mathbb{R}^{d_{\mathrm{model}}}$ denote the
cached final-layer residual for example $b$ at sequence position $j$ after
recurrent pass $t$. We apply the checkpoint's native final normalization and
output projection,
\[
    \ell^{(t)}_{b,j}
    =
    W_{\mathrm{out}}\,
    \mathrm{LN}_{\mathrm{out}}
    \!\left(h^{(t)}_{b,j}\right),
    \qquad
    p^{(t)}_{b,j}
    =
    \operatorname{softmax}\!\left(\ell^{(t)}_{b,j}\right),
\]
where $\mathrm{LN}_{\mathrm{out}}$ and $W_{\mathrm{out}}$ are exactly those
used by the corresponding model at its normal output readout. Thus, no
auxiliary probe is trained: the analysis asks whether an intermediate state
is already aligned with the model's learned output space.

Because each ground-truth state in our controlled tasks corresponds to a
vocabulary item, we can directly measure its readout probability. For a
target state $y_b$, we report either its batch-mean probability,
\[
    P^{(t)}_j(y)
    =
    \frac{1}{B}
    \sum_{b=1}^{B}
    p^{(t)}_{b,j}(y_b),
\]
or the corresponding vocabulary-wide argmax accuracy,
\[
    A^{(t)}_j(y)
    =
    \frac{1}{B}
    \sum_{b=1}^{B}
    \mathbf{1}
    \left[
        \arg\max_v p^{(t)}_{b,j}(v)=y_b
    \right].
\]
When we report a \emph{peak batch-mean probability} for DR-Loop, we first average the
target probability across examples at each recurrent pass and then maximize
over passes; this is distinct from averaging per-example maxima.

Logit-lens decodability is representational rather than causal evidence:
a high score shows that the residual is aligned with a task value under the
model's own readout, but does not by itself establish that the model uses
that information in subsequent computation. We therefore pair these
measurements with activation interventions that test the causal role of the
corresponding residual states.

\subsection{Activation Patching and State-Swap Conventions}
\label{app:patching-conventions}

We use activation patching \citep{meng2022locating, wang2022interpretability} to test whether a particular internal state
causally influences subsequent computation. Conceptually, we run a
\emph{recipient} computation normally except at a specified recurrent pass,
sequence position, and residual site, where we replace its activation with
one obtained from a \emph{donor} computation. We then allow the modified
forward computation to continue and measure whether downstream internal
states or predictions change in the counterfactual direction.

Let
\[
    h^{(t,s)}_{b,j}\in\mathbb{R}^{d_{\mathrm{model}}}
\]
denote the residual state for example $b$, sequence position $j$, recurrent
pass $t$, and residual site $s$, where $s$ specifies a location within the
shared Transformer stack (e.g., the input to block $0$ or the output of a
particular block). A generic patch replaces the recipient activation by
\[
    \widetilde h^{(t,s)}_{b,j}
    \leftarrow
    h^{(\tau,s)}_{b',j'},
\]
where $b',j',\tau$ specify the donor example, position, and recurrent pass.
Unless otherwise stated, all unpatched activations follow the recipient's
normal forward computation.

We refer to whole-residual replacements of this form as \emph{state swaps}.
Different experiments vary which axis is changed. A temporal state swap uses
the same example and token position but a donor from another recurrent pass
($b'=b$, $j'=j$, $\tau\neq t$); a cross-example swap uses another example,
typically at the same recurrent pass; and a cross-position transplant moves
a residual state to a different sequence position ($j'\neq j$). Some
experiments apply a replacement at a single residual site, whereas others
overwrite the corresponding state at multiple sites within one recurrent
pass or sustain the intervention across subsequent passes. We specify the
duration and patched sites for each experiment.

The key diagnostic is whether the intervention produces the corresponding
counterfactual computation rather than merely degrading the model. For
example, if a donor residual carries entity $e$ while the recipient retains
operator $r$, the predicted counterfactual next state is $r(e)$. We therefore
compare the intervened computation against this counterfactual target,
alongside the recipient's native target where appropriate.

State swaps provide causal evidence that the patched residual state
influences downstream computation, but a whole-residual replacement may
transfer multiple properties simultaneously, including semantic content,
computational status, and other contextual information. We therefore use
targeted counterfactuals, control interventions, and complementary
representational analyses to distinguish these possibilities where needed.

\section{Detailed Mechanistic Analysis on MR-Loop}
\label[appendix]{app:mrloop-mechanism} 

\subsection{Attention Analysis}
\label[appendix]{app:mrloop-kg-attention}

We inspect attention matrices across recurrent passes for the
knowledge-graph MR-Loop checkpoint analyzed in Section~\ref{sec:RQ1}. Details of the attention-analysis conventions used in this study are provided in \Cref{app:attention_conventions}.  Two recurring patterns are especially salient. 

\begin{figure}[htbp]
    \centering
    \includegraphics[width=1\linewidth]{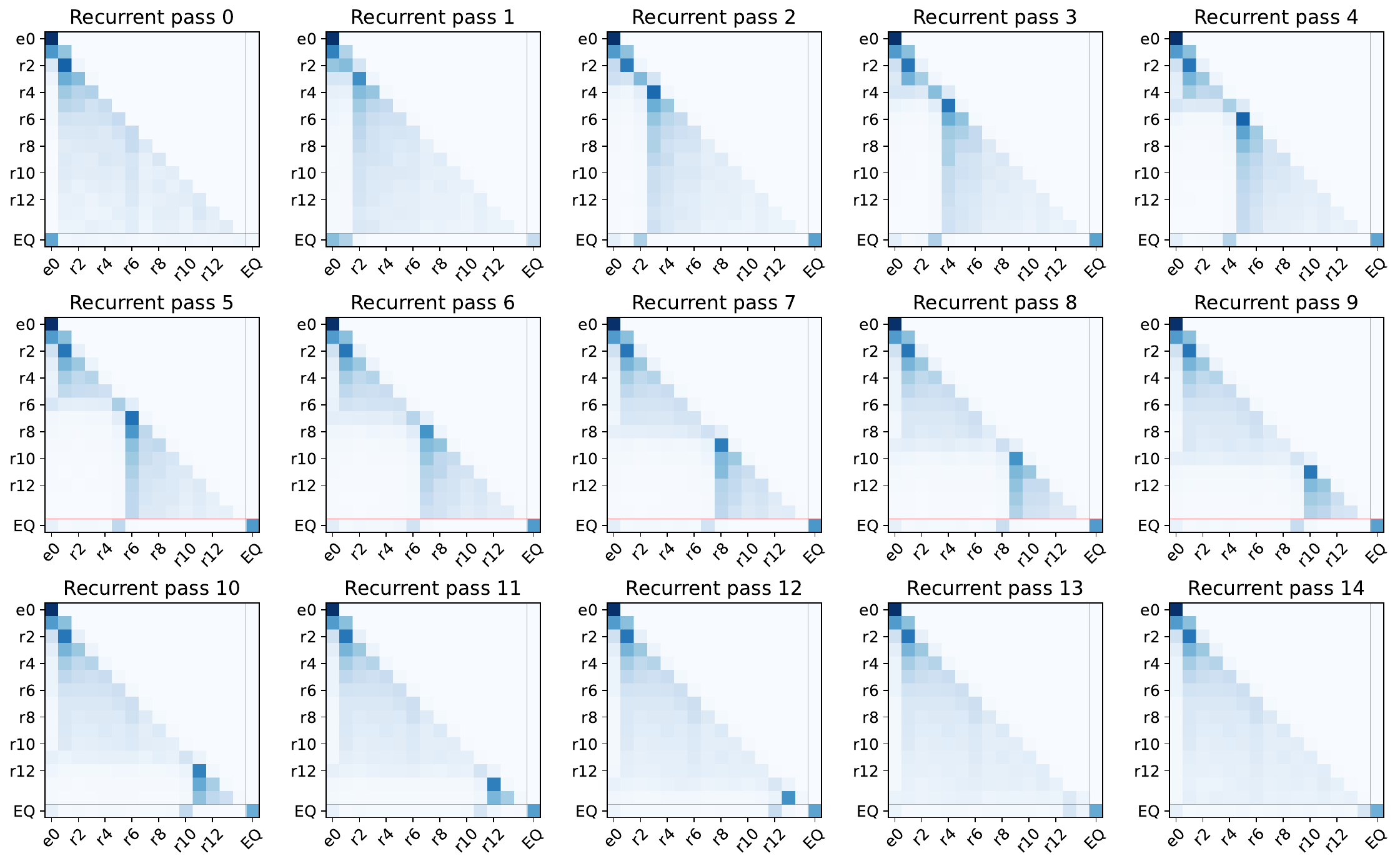}
    \caption{\textbf{Traveling previous-token attention in MR-Loop.}
Head L0H1 implements a recurrently advancing local attention pattern: at the recurrent pass (loop) $t$, $r_{t+2}$ attends strongly to its predecessor $r_{t+1}$. As recurrence proceeds, the attention peak shifts forward by one relation per loop and continues beyond the maximum training length $n=12$, suggesting a shared local mechanism for propagating computational progress to unseen relation positions.}
    \label{fig:mrloop_l3h2_kg_l0h1}
\end{figure}

\paragraph{Traveling previous-token attention.}
\Cref{fig:mrloop_l3h2_kg_l0h1} shows a head whose strongest local
interaction advances through successive relation positions as
recurrence proceeds. At each pass, a relation position attends strongly
to its immediately preceding relation position; the location of this
interaction shifts forward by approximately one relation per recurrent
pass. Importantly, the same pattern continues at relation indices beyond
those encountered during training.

\begin{figure}[htbp]
    \centering
    \includegraphics[width=1\linewidth]{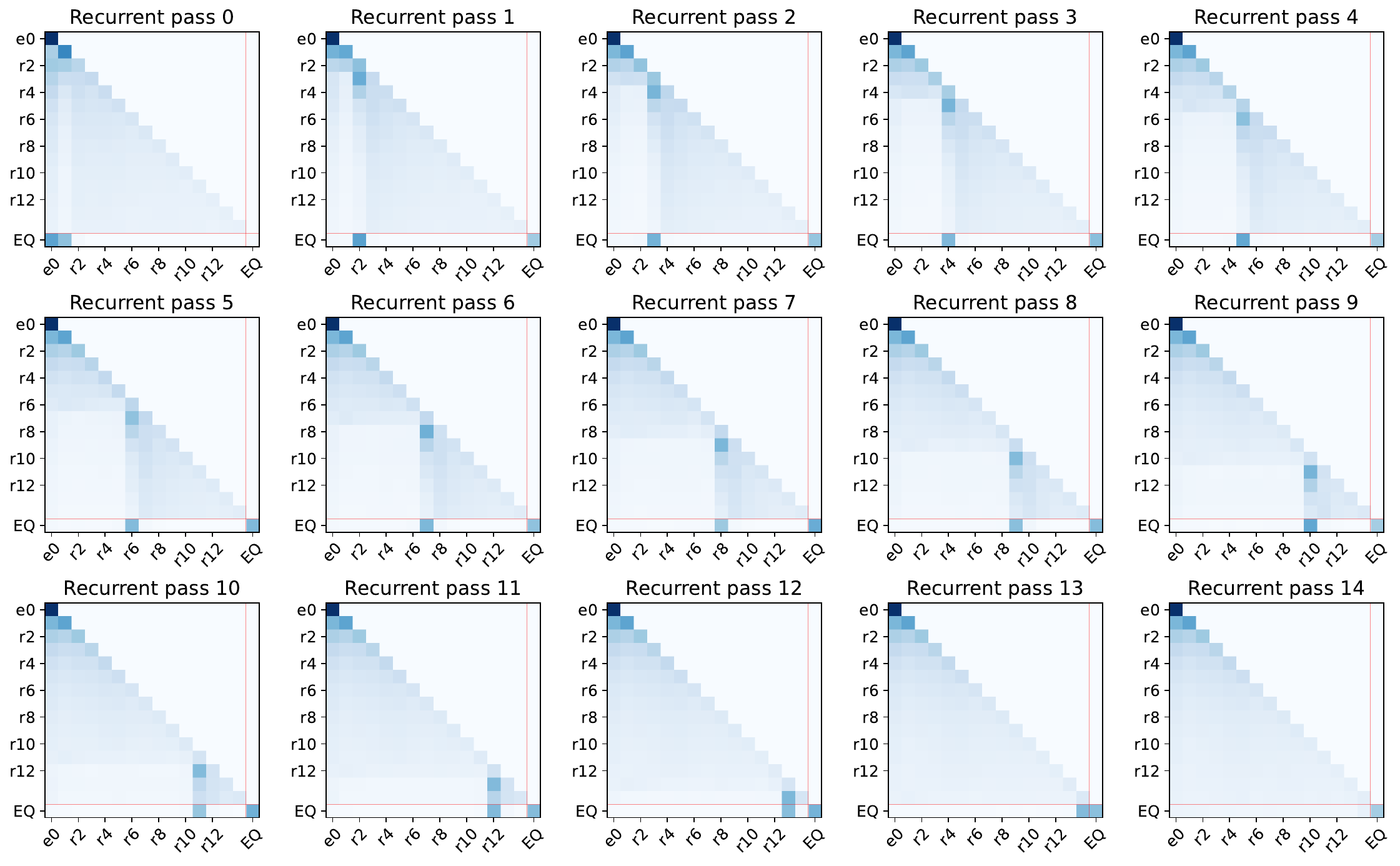}
    \caption{\textbf{Fixed-query tracking attention in MR-Loop.} Head L2H0 maintains its query at the \texttt{[EQ]} readout position while the attended relation advances across recurrent steps: at recurrent pass (loop) $t$, \texttt{[EQ]} attends strongly to $r_{t+1}$. This tracking pattern progresses by one relation per loop and extends beyond the maximum training length $n=12$, providing a mechanism for sequentially routing relation information into the evolving readout state.}
    \label{fig:mrloop_l3h2_kg_l2h1}
\end{figure}

\paragraph{Fixed-query tracking attention.}
Figure~\ref{fig:mrloop_l3h2_kg_l2h1} shows a complementary head whose query
remains at the fixed \texttt{[EQ]} readout position while its attended
relation advances through the sequence. Thus, recurrent progress occurs
along the key axis while the query position remains fixed. This pattern
is consistent with routing successive relation information into the
evolving state stored at \texttt{[EQ]}.

\subsection{MR-Loop intermediate value tracking}
\label[appendix]{app:mrloop-kg-intermediate}

\paragraph{Setup.}
We use the logit-lens procedure defined in
\Cref{app:logit-lens} to determine where MR-Loop represents the evolving
intermediate entity. We evaluate 32 randomly sampled length-$n=15$
knowledge-graph walks,
\[
    [e_0,r_1,\ldots,r_{n-1},\texttt{[EQ]}],
\]
which extend beyond the maximum training length $n=12$.
The ground-truth trajectory is
\[
    e_k = r_k(e_{k-1}), \qquad k=1,\ldots,n-1.
\]
Recurrent passes are zero-indexed. For this checkpoint, the
\texttt{[EQ]} readout after pass $t$ aligns with the look-ahead state
$e_{t+1}$, with the target capped at the final entity once the walk is
complete.

\begin{figure}[htbp]
    \centering
    \includegraphics[width=1\linewidth]{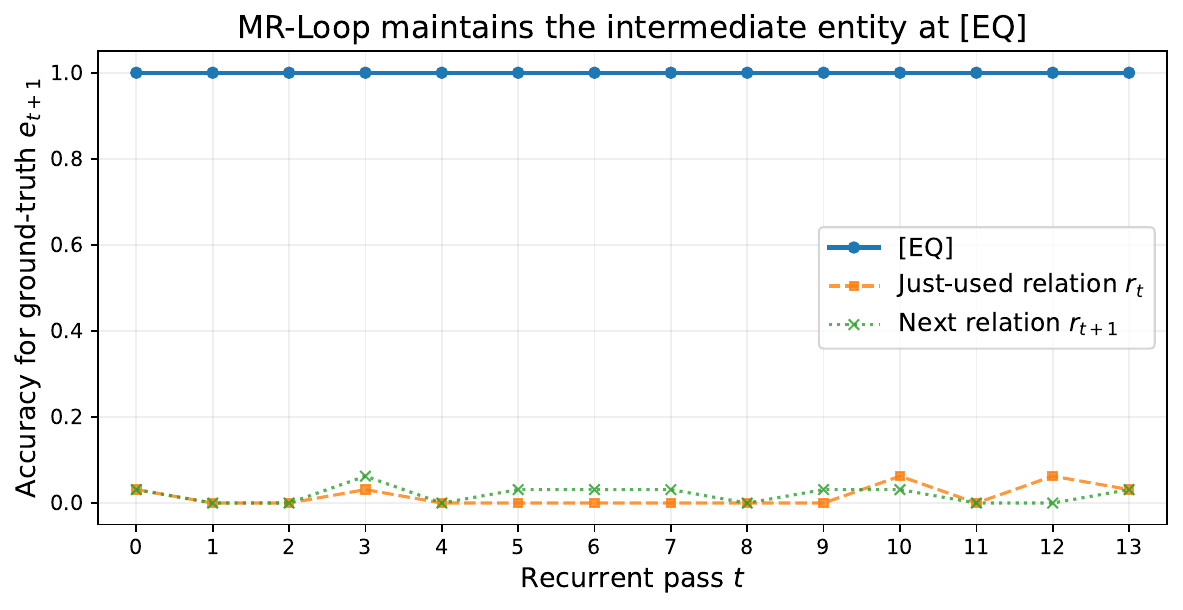}
    \caption{\textbf{Intermediate-state tracking in MR-Loop.}
After each loop $t$, we decode the last-layer residual at $[\mathrm{EQ}]$, the just-used key $r_t$, and the next unread key $r_{t+1}$, and measure whether the argmax recovers the next walk entity $e_{t+1}$. On length-$15$ walks, beyond the maximum training length $n=12$, $[\mathrm{EQ}]$ tracks $e_{t+1}$ with perfect accuracy, while predictions at $r_t$ and $r_{t+1}$ remain low. This shows that the evolving intermediate state is maintained as a look-ahead representation at the fixed $[\mathrm{EQ}]$ readout position rather than on the relation tokens.}
    \label{fig:mrloop-kg-intermediate-track}
\end{figure}

After every recurrent pass, we decode the final-layer residual at
\texttt{[EQ]} and at the two neighboring relation-token positions
$r_t$ and $r_{t+1}$ used in the analysis of
\Cref{fig:mrloop-kg-intermediate-track}. We then measure vocabulary-wide
argmax accuracy with respect to the same target $e_{t+1}$. This comparison
tests whether the evolving task state is preferentially localized at the
fixed answer readout or remains represented at the relation-token sites
participating in the advancing attention pattern.

\paragraph{Results.}
As shown in \Cref{fig:mrloop-kg-intermediate-track}, the intermediate entity
is directly decodable from \texttt{[EQ]} throughout the recurrent
computation. The \texttt{[EQ]} readout predicts $e_{t+1}$ correctly on
the evaluated passes with accuracy $1.00$. In contrast, decoding accuracy at
$r_t$ and $r_{t+1}$ remains close to zero.

The same alignment persists beyond the training horizon: at recurrent
steps corresponding to relation indices not encountered during training,
the correct intermediate remains perfectly decodable from
\texttt{[EQ]}. Thus, while attention advances through successive relation
positions, the task state itself remains localized at a fixed readout
position. Combined with the attention analysis in \Cref{app:mrloop-kg-attention},
this supports a recurrent organization in which relation selection moves
through the input while \texttt{[EQ]} carries the evolving intermediate
entity. The state-swap experiments in \Cref{app:mrloop-kg-state-swaps} then
test this interpretation causally. 

\subsection{MR-Loop State-swap experiments}
\label[appendix]{app:mrloop-kg-state-swaps}

We use the activation-patching and state-swap framework defined in
\Cref{app:patching-conventions} to test the proposed separation between
relation selection and the evolving entity stored at \texttt{[EQ]}.

\paragraph{Setup and evaluation.}
We sample 64 knowledge-graph walks and construct matched inputs of different lengths by truncating each walk and appending \texttt{[EQ]}. Length
$n$ counts the tokens before \texttt{[EQ]}: one starting
entity followed by $n-1$ relations. Training covers $n\leq 12$, corresponding to relation
indices $k\leq 11$.  

For this checkpoint, recurrent pass $t$ is aligned with the transition
\[
    e_{t+1}=T(e_t,r_{t+1}).
\]
Each intervention is applied during a specified pass $t$. Unless otherwise
stated, all unpatched states follow the recipient computation. We evaluate
the vocabulary-wide argmax at \texttt{[EQ]} immediately after the
intervention pass. Counterfactual accuracy is computed only on examples for
which the counterfactual target differs from the recipient's native next
entity. For attention measurements, we average the \texttt{[EQ]} attention
row over all 64 examples and take its argmax over pre-\texttt{[EQ]} input
positions, excluding \texttt{[EQ]} self-attention.

\paragraph{Temporal state swaps test relation selection.}
To test whether relation-token residual states determine which relation is
selected, we perform same-example temporal state swaps. The example and token
position are held fixed, while the residual state of that token is replaced
by its cached state from a different recurrent pass. At each patched residual
site, we use the activation from the corresponding site in the donor pass.
The incoming \texttt{[EQ]} state and all input tokens remain unchanged.

At recipient loop $t$, the unmodified model normally selects relation
$r_{t+1}$ and computes $T(e_t,r_{t+1})$. We test whether changing only the
recurrent state of a relation token can reroute this selection backward or
forward.

For \emph{backward rerouting}, we replace the residual state of the previous
relation token $r_t$ with its state from loop $t-1$, when $r_t$ was normally
selected. If this restores its selectability, the model should select $r_t$
again and produce the counterfactual output
\[
    T(e_t,r_t).
\]

For \emph{forward rerouting}, we replace the currently scheduled relation
token $r_{t+1}$ with its own state from a later loop $t+d$, with
$d\in\{1,2,3,4\}$. The token position remains fixed at $t+1$; only its
recurrent state is changed. We test whether this later state suppresses
selection of $r_{t+1}$, causing the model to advance to $r_{t+2}$ and produce
\[
    T(e_t,r_{t+2}).
\]

For example, at $t=5$, the native computation selects $r_6$. We replace the
residual state of $r_6$ with its cached state from loop $6$, $7$, $8$, or $9$.
In every condition, the batch-mean attention pointer shifts to $r_7$, and we
therefore evaluate the counterfactual target $T(e_5,r_7)$. Notably, although
the unmodified model at loop $9$ selects $r_{10}$, transplanting $r_6$'s
loop-$9$ state into the loop-$5$ computation still causes a one-relation
shift from $r_6$ to $r_7$. Thus, across the tested donor ages, the later
state of the scheduled relation consistently induces a one-relation skip. 

At $(n,t)=(12,5)$, backward rerouting achieves $100\%$
counterfactual accuracy ($58/58$), and forward rerouting achieves
$100\%$ ($58/58$) for each of the four donor ages. At
$(n,t)=(20,13)$, beyond the trained hop range, backward rerouting
achieves $91.5\%$ ($54/59$), while forward rerouting achieves
$93.1\%$ ($54/58$) for every donor age. The batch-mean attention
pointers of L1H1, L2H0, and L2H1 move from the scheduled key
$t+1$ to key $t$ under backward rerouting and to key $t+2$ under
all four forward interventions. Here, pointers are computed over
pre-\texttt{[EQ]} input positions, excluding \texttt{[EQ]}
self-attention.

In the unmodified donor runs, the corresponding pointers advance
to key $t+d+1$ as the donor loop increases. After transplantation,
however, the recipient pointers remain at key $t+2$. Thus, across
the tested donor ages, later states of the scheduled relation
consistently induce a one-relation skip. These results are
consistent with later token states suppressing selection of that
relation. Without intervention, accuracy on each counterfactual target is
$0\%$ on its evaluation subset. Unintervened native-hop accuracy
is $100\%$ ($64/64$) at $(12,5)$ and $93.8\%$ ($60/64$) at
$(20,13)$.

Together, these interventions provide causal evidence that
relation-token residual states help determine which relation is
applied to the entity at \texttt{[EQ]}. Altering these states changes
both attention routing and the predicted entity in the expected
counterfactual direction, including beyond the trained hop range.
The observed state differences therefore have a functional role
in relation selection, supporting their interpretation as
computational progress information.

\paragraph{Intervening on the intermediate entity.} 
We next test whether the residual at \texttt{[EQ]} supplies the entity
operand for the next relation application.  For recipient example $i$, we replace only the \texttt{[EQ]} residual
at the input to block~0 with the same-loop residual from donor
example $j$, selected by cyclically shifting the batch. All
relation-token states remain those of the recipient, and the model then
continues its forward computation normally.

If the transplanted \texttt{[EQ]} state carries the donor intermediate
entity $e_t^j$ while the recipient continues to select relation
$r_{t+1}^i$, the predicted counterfactual output is
\[
    T(e_t^j, r_{t+1}^i).
\]
On length-$20$ inputs, counterfactual accuracy is $100\%$ at hop
$k=t+1=5$, $93.8\%$ at the training boundary $k=11$, and $95.2\%$
at the unseen hop index $k=14$. At each of these indices, the
batch-mean L1H1 attention argmax remains at the native relation key,
and the recipient's native next entity is predicted on $0\%$ of
the counterfactual evaluation examples. Thus, the swapped
\texttt{[EQ]} state causally supplies the entity used in the next
relation application, including beyond the training horizon.

\section{MR-Loop: A transferable live--consumed state cue}
\label[appendix]{app:mrloop-kg-length-mechanism}

\paragraph{Experimental setup.}
We analyze the same MR-loop (analyzed throughout \Cref{sec:RQ1}) using $64$ sampled knowledge-graph
walks, with shorter inputs constructed as prefixes of the same walks.
Training covers lengths $n \leq 12$, corresponding to relation
indices $k \leq 11$. For intermediate loops $t \geq 1$, the proposed
computation updates the entity at \texttt{[EQ]} according to
$e_{t+1}=T(e_t,r_{t+1})$, where $T$ is the graph transition function.
Hop accuracy measures whether the argmax prediction at \texttt{[EQ]}
after loop $t$ equals the specified target.

\paragraph{Estimating the live--consumed direction.} 
For relation key $k$, we define its \textbf{\emph{live}} state as its
residual during its normal fetch loop $k-1$, and its
\textbf{\emph{consumed}} state as its residual at loop $k+1$.
At each residual site $s$, we compute
\[
\Delta h_{s,i,k}
=
h^{\mathrm{live}}_{s,i,k}
-
h^{\mathrm{consumed}}_{s,i,k},
\qquad
v_s
=
\mathbb{E}_{i,k}\!\left[\|\Delta h_{s,i,k}\|_2\right]
\frac{
\mathbb{E}_{i,k}[\Delta h_{s,i,k}]
}{
\left\|\mathbb{E}_{i,k}[\Delta h_{s,i,k}]\right\|_2
}.
\] 
Here, \(s\) indexes the residual recording site within a loop, \(i\) indexes the sampled walk, and \(k\) indexes the relation-token position. For our three-block Transformer with
two attention heads per block, we use four recording sites:
immediately before block~0, immediately after block~0,
immediately after block~1, and immediately after block~2.  For each fixed site \(s\), we average over all 64 walks and positions \(k=5,\ldots,11\) to obtain one shared vector \(v_s\), using their length-$30$ caches.
Thus, $v_s$ follows the mean \textbf{live-minus-consumed difference},
with magnitude matched to the mean individual difference norm.
We estimate separate directions at the input to block~0 and
after each block. Transfer is evaluated at additional positions
on the same walks.

\paragraph{Causal intervention and transfer beyond training.}

At the loop that normally fetches relation key $k$, namely
$t=k-1$, we intervene on that key's residual at the input to
block~0 and after every block. At each site $s$, we replace
the residual with either $h_s^{\mathrm{consumed}}$, cached
from the same example and key position at the later loop
$k+1$, or $h_s^{\mathrm{consumed}}+v_s$.
The input tokens and incoming \texttt{[EQ]} residual remain
unchanged. The first intervention gives the scheduled relation token
the state it normally has two loops after its selection.
This tests whether that later state causes the model to
treat the relation as already consumed and skip it.
The second intervention adds the shared live-minus-consumed
direction $v_s$ to the same later state, testing whether
restoring this cue makes the key selectable again.

For example, on a length-$12$ input, loop $t=5$ normally
selects key $6$ to compute $T(e_5,r_6)$.
We replace key $6$'s residuals with its own cached residuals
from loop $7$, using the corresponding donor activation
at each site. This consumed-state replacement shifts the
batch-mean L1H1 attention pointer from key $6$ to key $7$,
consistent with skipping $r_6$ and applying $r_7$ to the
incoming entity $e_5$.
Adding $v_s$ to those consumed residuals returns the pointer
to key $6$ and restores correct-hop accuracy from $9.4\%$
to $100\%$.
Here, attention pointers are computed over input-token
positions, excluding \texttt{[EQ]}.

At $(n,k)=(12,6)$, 
replacing the scheduled key's live residual
$h_s^{\mathrm{live}}$ with its consumed residual
$h_s^{\mathrm{consumed}}$ at every intervention site $s$
reduces accuracy on this target from $100\%$ to $9.4\%$.
Using $h_s^{\mathrm{consumed}}+v_s$ as the replacement
restores accuracy on the same target to $100\%$. 
The same directions transfer beyond the training boundary. At $(n,k)=(20,14)$, loop $t=13$ normally selects key $14$
to compute $e_{14}=T(e_{13},r_{14})$.
Replacing that key's residual with
$h_s^{\mathrm{consumed}}$ at every intervention site $s$
reduces accuracy on this target to $9.4\%$.
Using $h_s^{\mathrm{consumed}}+v_s$ as the replacement
increases the batch-mean L1H1 attention mass from
\texttt{[EQ]} to key $14$ from $0.01$ to $0.47$ and
restores accuracy to $93.8\%$, equal to the accuracy
of the unintervened run. On the same length-$20$ inputs, we repeat this intervention
at key $17$ during loop $t=16$, with target
$e_{17}=T(e_{16},r_{17})$.
Replacing the key's residuals with 
$h_s^{\mathrm{consumed}}$ gives $9.4\%$ accuracy,
whereas replacing them with
$h_s^{\mathrm{consumed}}+v_s$ gives $85.9\%$.
The unintervened run achieves $81.2\%$ on this target.
The same shared directions $v_s$ are used at both positions.

Two controls establish the specificity of this effect.
Subtracting $v_s$ from a live state reproduces the disruption
caused by replacing it with a consumed state.
Adding a random direction of the same norm to the consumed
state leaves accuracy at $9.4\%$ in both the $(12,6)$ and
$(20,14)$ settings.
These interventions show that the estimated directions
causally affect relation selection and recover the correct
entity transition at all the positions.

\paragraph{Key states control computational progress.}
Complementary interventions locate the progress signal in
relation-key states. For matched loops and shared prefixes,
residuals at the loop input and final block output agree
across $n \in \{12,20,30\}$ to within $2.7 \times 10^{-5}$.
Freezing every relation key at its loop-$t_0$ state, with
replacements at the loop input and after every block,
prevents selection from advancing. On $n=20$, freezing from
$t_0=12$ holds the L1H1 pointer at key $13$ throughout the
remaining simulated loops.

Conversely, replacing the \texttt{[EQ]} residual or attention
queries with those from another loop does not systematically
transfer the donor's selection position. Across $32$
interventions, L1H1 selects the donor-loop position on only
$0.9\%$ of example--intervention pairs.
Some swaps impair routing, indicating that \texttt{[EQ]}
can affect routing quality even though key states control
which relation is selected.

\paragraph{Local interactions support advancement.}
We remove attention edges from relation position $q$ to $q-1$
across all heads and layers, renormalizing the affected rows
while leaving the \texttt{[EQ]} attention row untouched.
On $n=20$, restricting this intervention to $q>11$ causes
the L1H1 pointer to first deviate by more than one position
from the expected key at loop $t=11$.
Mean intermediate-state accuracy over $11 \leq t \leq 18$
falls from $84.0\%$ to $0.6\%$.
Earlier computation is largely preserved: mean accuracy
over $t \leq 10$ is $94.0\%$, compared with $97.9\%$ natively.
Applying the same ablation to $q \leq 11$ produces pointer
failure at $t=1$ under the same criterion.
These results establish a necessary role for interactions
between adjacent relation tokens in continued advancement
beyond the training range.

\paragraph{The selected relation acts on the entity at \texttt{[EQ]}.}
We replace the recipient's \texttt{[EQ]} residual at the input
to block~0 with another example's same-loop residual, leaving
relation-key states unchanged. The counterfactual target is
$T(e_t^j,r_{t+1}^i)$ for donor $j$ and recipient $i$.
On $n=20$, counterfactual accuracy is $93.8\%$ at the training
boundary $k=t+1=11$ and $95.2\%$ at $k=14$, evaluated on
examples where this target differs from the recipient's
native next entity. At both indices, the batch-mean L1H1
pointer remains at the recipient's next relation, and the
native next entity is predicted on $0\%$ of these examples.
Thus, the intermediate entity at \texttt{[EQ]} causally
supplies the operand for the selected relation, including
at unseen depths.

\paragraph{Summary.}
Together, these experiments support a mechanism in which a
transferable key-state cue guides relation selection,
adjacent-token interactions support its advancement, and
\texttt{[EQ]} supplies the entity for composition.
The observed extrapolation has a finite operating range:
final-answer accuracy with $R=n$ loops is $70.3\%$ at $n=20$
and $1.6\%$ at $n=30$, and the key-state cue weakens at
larger hop indices.

\section{Detailed Mechanistic Analysis on DR-Loop}
\label[appendix]{app:drloop_detailed_mechanism} 

\subsection{Dr-Loop attention analysis}
\label[appendix]{app:drloop_detailed_mechanism_attention} 

We inspect attention matrices across recurrent passes for the
knowledge-graph DR-Loop checkpoint analyzed in
\Cref{sec:RQ1}. Attention-map conventions are given in
\Cref{app:attention_conventions}. Unlike MR-Loop, where recurrent
progress is expressed through a fixed \texttt{[EQ]} query and
advancing relation selection, DR-Loop exhibits localized attention
patterns whose active region itself moves toward later sequence
positions as recurrence proceeds.

\begin{figure}[htbp]
    \centering
    \includegraphics[width=1\linewidth]{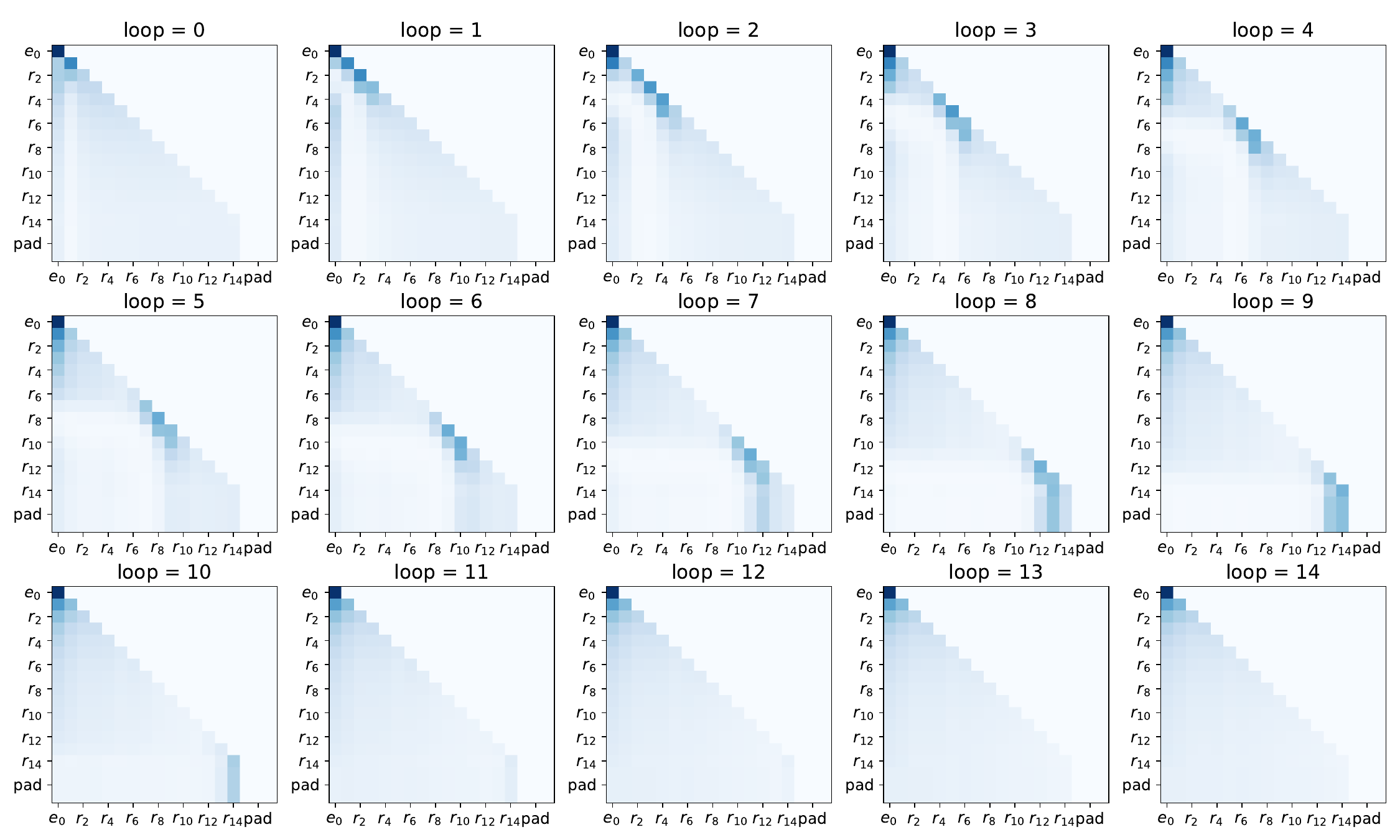}
\caption{
\textbf{Traveling self-attention head in DR-Loop.}
A localized block of elevated attention advances toward later sequence positions along the main diagonal across recurrent iterations. The highlighted region spans approximately two to three neighboring positions and moves forward by roughly one to two positions per loop, indicating that the head sequentially tracks the advancing computational frontier. The pattern continues through relation positions beyond the training horizon until reaching the end of the sequence, after which the localized late-position attention disappears in subsequent iterations.
}
    \label{fig:drloop-self-attn}
\end{figure}

\paragraph{Traveling self-attention.}
\Cref{fig:drloop-self-attn} shows a representative head with a
localized region of elevated attention near the main diagonal,
corresponding to queries attending strongly to their own positions.
The active region spans approximately two to three neighboring
positions and advances toward later relation positions by roughly
one to two positions per recurrent pass. This progression continues
to relation positions beyond those encountered during training.
After the active region reaches the end of the relation sequence,
the localized late-position pattern weakens in subsequent passes.

\begin{figure}[htbp]
    \centering
    \includegraphics[width=1\linewidth]{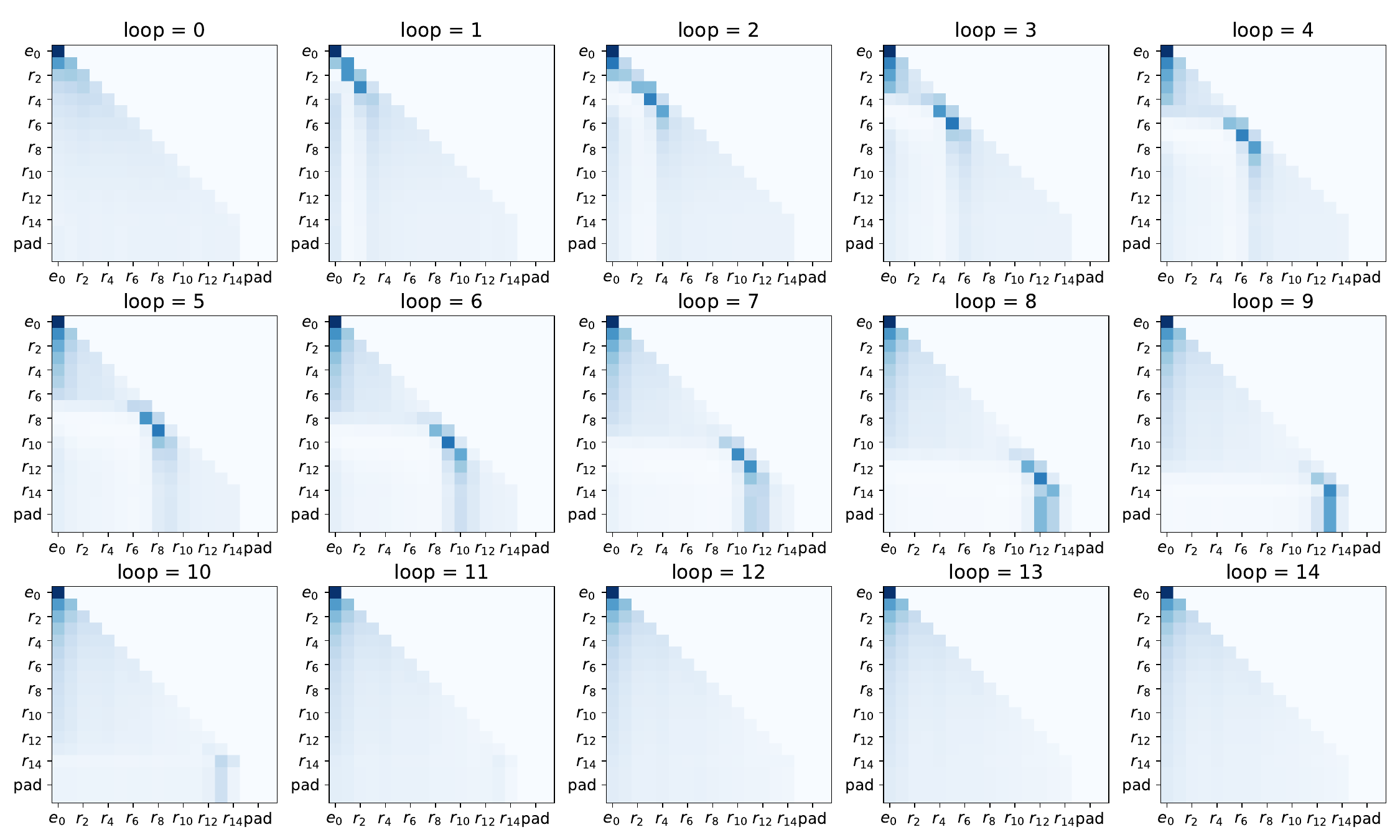}
    \caption{
\textbf{Traveling previous-position attention head in DR-Loop.}
A localized block of elevated attention advances toward later sequence positions along the first subdiagonal across recurrent iterations, corresponding to queries attending strongly to the immediately preceding position. The highlighted region spans approximately two to three positions and moves forward by roughly one to two positions per loop, qualitatively tracking the advancing intermediate-state frontier. This traveling pattern extends to relation positions beyond those encountered during training and weakens after reaching the end of the sequence.
}
    \label{fig:drloop-prev-attn}
\end{figure}

\paragraph{Traveling previous-position attention.}
\Cref{fig:drloop-prev-attn} shows a complementary head whose localized
attention region travels along the first subdiagonal. Here, queries
near the advancing region attend strongly to the immediately preceding
sequence position. As recurrence proceeds, this region likewise moves
forward by roughly one to two positions per pass and extends beyond the
training horizon before weakening near the end of the sequence.

Together, these patterns indicate that DR-Loop organizes information
routing around an advancing local region rather than a fixed readout
position. The motion of this attention region qualitatively tracks the
advancing intermediate-state frontier identified by the logit-lens
analysis in \Cref{app:drloop-logit-lens}, motivating the state-propagation
mechanism tested causally in \Cref{app:drloop-state-swap}.

\subsection{Dr-Loop logit-lens analysis}
\label[appendix]{app:drloop-logit-lens}

\paragraph{Setup.}
We use the logit-lens procedure defined in
\Cref{app:logit-lens} to determine where intermediate entities become
represented as DR-Loop recurrent computation progresses. We analyze the
same knowledge-graph checkpoint as in the main text, trained on inputs of length at most $12$, and evaluate $64$ queries of length $15$ using $15$ recurrent passes.

For each example, position $0$ contains the starting entity $e_0$, while
position $k\geq 1$ contains relation $r_k$. The ground-truth trajectory is
\[
    e_k = r_k(e_{k-1}), \qquad k=1,\ldots,14.
\]
After each recurrent pass $t$, we apply the model's native final
normalization and unembedding to the final-layer residual at every
sequence position. For each intermediate entity $e_k$, we measure its
batch-mean logit-lens probability at every residual position.

\begin{figure}[htbp]
    \centering
    \includegraphics[width=1\linewidth]{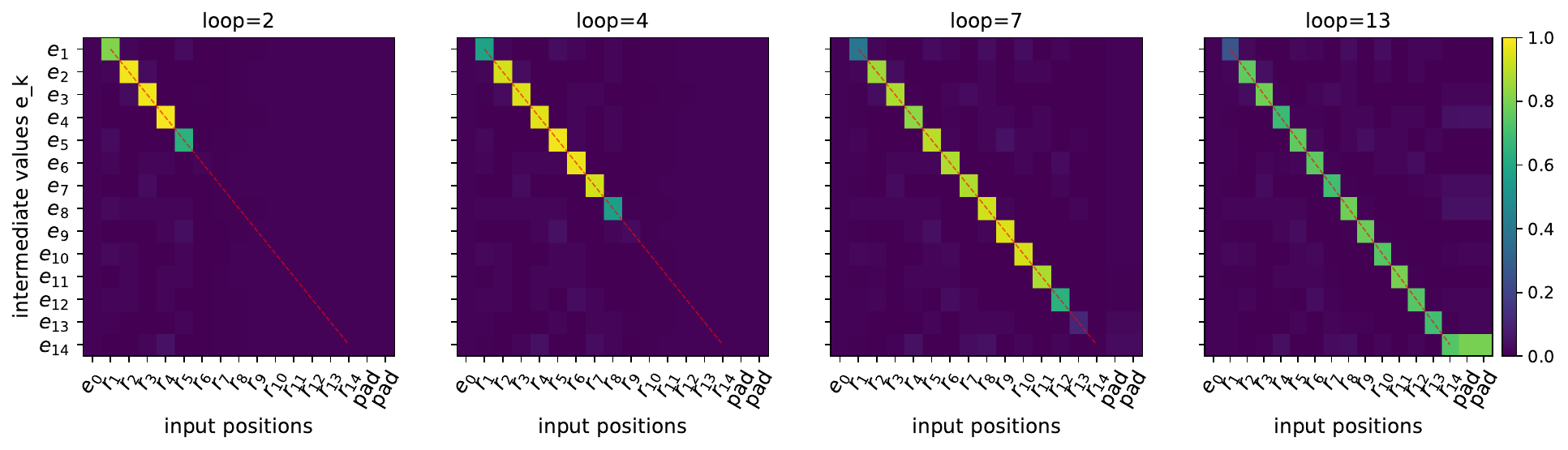}
\caption{
\textbf{Intermediate-state propagation in DR-Loop.}
Logit-lens probability assigned to each ground-truth intermediate entity $e_k$ across residual positions at selected recurrent iterations. The dashed diagonal marks the expected alignment between intermediate entity $e_k$ and relation position $k$. As recurrence proceeds, a high-probability band advances toward later positions, showing that successive intermediate entities become decodable at their corresponding relation sites. This frontier continues beyond the maximum training depth of 11 hops: $e_{12}$ and $e_{13}$ become strongly decodable at unseen relation positions. The progression of this representational frontier qualitatively tracks the traveling attention patterns in \Cref{fig:drloop-prev-attn}, providing evidence that DR-Loop propagates intermediate states through the sequence during length generalization.
}
    \label{fig:drloop_l3h2_kg_logitlens}
\end{figure}

Importantly, we do not assume a one-to-one correspondence between
recurrent pass $t$ and task hop $k$. Instead, we test the
\emph{state-to-position} alignment predicted by the propagating-frontier
hypothesis: intermediate entity $e_k$ should become decodable at the
relation position $k$ as the computation reaches that part of the
sequence. Thus, the diagonal in \Cref{fig:drloop_l3h2_kg_logitlens} marks
the expected correspondence between entity index $k$ and relation
position $k$, rather than between hop index and recurrent-pass index.

\paragraph{Results.}
As shown in \Cref{fig:drloop_l3h2_kg_logitlens}, the region of strong
intermediate-state decodability advances toward later sequence positions
as recurrence proceeds. At early passes, only early intermediate entities
are strongly decodable at their corresponding relation positions. With
additional recurrence, this high-probability frontier moves progressively
toward later positions, closely paralleling the traveling attention
patterns in \Cref{app:drloop_detailed_mechanism_attention}.

The representational frontier also extends beyond the training depth.
For positions within the trained range, $1\leq k\leq 11$, the correct
entity $e_k$ reaches peak batch-mean probabilities of approximately
$0.95$--$1.00$. At unseen positions, $e_{12}$ and $e_{13}$ remain
strongly decodable, reaching approximately $0.90$--$0.92$. Because an
intermediate representation can weaken after it is formed, these summary
values report the maximum batch-mean probability over recurrent passes
at each position.

These results provide representational evidence for an advancing
intermediate-state frontier: rather than accumulating the evolving entity
at a fixed readout position as in MR-Loop, DR-Loop makes successive
entities decodable at progressively later relation positions. Together
with the traveling attention patterns in \Cref{app:drloop_detailed_mechanism_attention},
this motivates the recurrent state-propagation mechanism tested causally
in \Cref{app:drloop-state-swap}.

\subsection{DR-Loop: Causal tests of recurrent state propagation}
\label[appendix]{app:drloop-state-swap}

\paragraph{Protocol and measurements.}
We use the same checkpoint as in the main analysis, trained on inputs of length at most $12$, and evaluate 64 length-15 queries  using 15 recurrent
passes. Each recipient is paired with a donor that has a different
starting entity but identical relations. Because each relation is a
permutation, donor and recipient entities differ at every hop, providing
unambiguous counterfactual targets $e'_j=f_{r_j}(e'_{j-1})$.
Query generation and donor construction use seeds 42 and 7, respectively.

We transplant the donor residual at position $k-1$ into the recipient
before the first transformer layer
(\texttt{blocks.0.hook\_resid\_pre}), using donor activations from the
same loop index. Replacement is sustained through the final pass;
all other activations follow the recipient's forward computation.
The interventions producing $e_6$ and $e_{13}$ begin at zero-based loop
indices $2$ and $7$, respectively. These indices are selected by the
first post-loop batch-mean logit-lens probability of the preceding
entity reaching $0.5$. Patching occurs before the loop with that
index, rather than at the following boundary where the measured
post-loop state first becomes available.

After the final pass, we apply the final normalization and unembedding
to the last-layer residual at each relation position. We report mean
probabilities for the original and counterfactual entities and the
counterfactual vocabulary-wide argmax rate over all 64 examples,
without conditioning on native correctness. Final-answer accuracy is
measured separately at the last PAD query (position 17 after padding
to 18 tokens). Native final-answer accuracy is $85.9\%$.

\paragraph{Counterfactual continuation beyond the training depth.}
Transplanting position $5$ reduces the original-entity probability at
position $6$ from $0.720$ to $0.058$, while the counterfactual probability
reaches $0.582$, with an argmax rate of $58/64$ ($90.6\%$).
Under this same position-$5$ intervention, the counterfactual also appears beyond the training range: at position
$14$, its probability is $0.592$ and its argmax rate is $46/64$
($71.9\%$). The final PAD predicts the counterfactual answer in $47/64$
examples ($73.4\%$), compared with the original answer in $8/64$
($12.5\%$). Native probabilities for the donor entities at positions
$6$--$14$ are approximately $0.002$--$0.004$.

\paragraph{Direct intervention at an extrapolated hop.}
To test causal influence at a hop beyond the training range, we
transplant position $12$ and measure the computation of $e_{13}$.
The original-entity probability decreases from $0.745$ to $0.038$,
while the counterfactual probability reaches $0.320$, with an argmax
rate of $36/64$ ($56.25\%$). This distinguishes direct intervention
at an extrapolated hop from the downstream continuation of the
earlier position-$5$ intervention.

\paragraph{Controls and interpretation.}
At $k=13$, a donor with a different prefix constructed to preserve
$e_{12}$ retains an original-entity probability of $0.344$, compared
with $0.038$ for the state-different donor and $0.745$ without
intervention. Thus semantic agreement preserves more of the original
computation, although the transplant remains disruptive.
Relocating the donor's position-$12$ residual to position $11$ yields
only $0.059$ probability for the donor entity expected at position
$13$; relocating it to position $14$ leaves the original probability
at position $13$ unchanged at $0.745$, as expected under causal
attention. These comparisons support a targeted, position-dependent
effect of intermediate residuals. However, whole-residual patches
also transfer computational phase and other prefix information, and
direct long-range attention paths remain available. Taken together, these results provide causal evidence that intermediate
residual states influence subsequent relation composition and downstream
answers, including beyond the training depth, supporting recurrent
state propagation as a mechanism for DR-Loop's length generalization.


\section{DR-Loop: Evidence for a local composition rule reused across hop indices}
\label[appendix]{app:dr-loop-kg-phase-reuse}

We test whether an entity representation obtained at an early position
can support the relation application at a later, unseen hop index.
Three comparisons address this claim: transfer of a fixed donor and
direction across frontiers, interventions on the donor's computational
phase, and interventions on the receiving residual. These complement
the preceding attention, intermediate-decoding, and state-swap analyses.

\paragraph{Setup and measurements.}
We use the same three-layer, two-head-per-layer checkpoint, trained on
at most 11 hops, with no positional embeddings or input reinjection.
The analysis uses 64 knowledge-graph walks of 29 hops, generated with
seed 42, and matched 11-hop prefixes. We run each 29-hop walk for 30 recurrent passes and its matched 11-hop prefix for 12 passes, where each pass reapplies the same three-layer transformer stack to the current hidden states. Position 0 contains the starting
entity; position $j\geq1$ contains relation $r_j$ and is the site
probed for the intermediate entity $e_j$.

Let $p_{b,j}^{(t)}(e)$ denote the probability of entity $e$ in example
$b$ at position $j$, obtained by applying the final layer normalization
and unembedding to the final-layer residual after zero-based loop $t$.
We define the native frontier as
\begin{equation}
 F_t=\max\left\{j:\frac{1}{B}\sum_{b=1}^{B}
 p_{b,j}^{(t)}(e_{b,j})\geq0.5\right\},\qquad B=64.
\end{equation}
For an intervention during loop $t^\star$, we set the intervention
position to $F=F_{t^\star-1}$, the frontier measured in the unperturbed
run after the preceding loop. The threshold of $0.5$ is applied to the
probability of the correct entity averaged across all 64 examples,
yielding a single intervention position shared by the entire batch. 


\paragraph{Donor construction and intervention.}
We obtain donor representations from the model's unmodified
(``native'') run. For a source position $z$, let $\tau_z$ be the
first loop after which the batch-mean logit-lens probability of the
correct entity at that position reaches $0.5$. For each example $b$,
the \textbf{\emph{fresh donor}} is the residual vector at position $z$ after
the final transformer layer of loop $\tau_z$. The \textbf{\emph{aged donor}}
is the residual vector at the same position and
in the same example five loops later, capped at the last evaluated
loop, i.e. after loop $\min(\tau_z+5, R-1)$.

During intervention loop $t^\star$, we copy the selected donor vector
from each example to its frontier position $F$. We hold this vector
fixed at $F$ throughout that pass by overwriting the residual before
the first transformer layer and again after every layer. Other
positions compute normally in response to the modified state.
The intervention ends after this pass: from loop $t^\star+1$
onward, all positions evolve without further replacement.
Thus, the intervention replaces the entire residual vector at one
position for one recurrent pass.


\paragraph{Counterfactual target and score.}
The donor represents an entity from source position $z$, whereas
the next relation in the recipient computation is $r_{b,F+1}$.
If the model applies this relation to the transplanted entity,
the resulting counterfactual entity should be
\[
  \widetilde e_{b,F+1}
  = f_{r_{b,F+1}}(e_{b,z}).
\]
We measure the logit-lens probability of this entity at position
$F+1$ after each loop, starting with the intervention loop.
At each loop, we first average this probability over the $B$
examples; we then take the largest of these batch means:
\begin{equation}
 S_{F,z}
 = \max_{t^\star \leq t < R}
   \left[
     \frac{1}{B}\sum_{b=1}^{B}
     p_{b,F+1}^{(t)}(\widetilde e_{b,F+1})
   \right].
\end{equation}
Here, $R$ is the total number of recurrent passes.
This score measures the strongest batch-level decoding of the
counterfactual next-hop entity during the remaining computation.
It is neither final-answer accuracy nor an average of
example-specific peak probabilities.

To test whether the counterfactual computation continues, we also
measure the probability at position $F+2$ of
\[
  \widetilde e_{b,F+2}
  = f_{r_{b,F+2}}(\widetilde e_{b,F+1}).
\]
We apply the same averaging-then-maximization procedure separately
to this continuation target and to each comparison label.
Their reported maxima may therefore occur at different loops.

\paragraph{A fixed intervention transfers across hop indices.}
We construct a shared intervention vector from paired fresh and aged
residuals. With $\delta_{b,z}=h^{\mathrm{fresh}}_{b,z}
-h^{\mathrm{aged}}_{b,z}$, define
\begin{equation}
 \overline\delta_z=\frac{1}{B}\sum_b\delta_{b,z},
 \qquad s_z=\frac{1}{B}\sum_b\|\delta_{b,z}\|_2,
 \qquad v_z=s_z\frac{\overline\delta_z}
                         {\|\overline\delta_z\|_2}.
\end{equation}
Thus the direction follows the batch-mean difference, with magnitude
set by the mean paired distance.

We use the same source-$2$ donor and the same $v_2$ at every tested
frontier in the 29-hop walks, without estimating a new direction
at later positions. The intervention loops are $t^\star=3,5,7,8,10,12$
for $F=5,8,11,12,14,16$, respectively.
Table~\ref{tab:kg-phase-transfer} compares the aged donor, fresh donor,
and adjusted donor $h^{\mathrm{aged}}_{b,2}+v_2$.
For the adjusted donor, we also compare the recipient-relation target
$f_{r_{b,F+1}}(e_{b,2})$ with the donor's original next entity
$e_{b,3}=f_{r_{b,3}}(e_{b,2})$.

\begin{table}[t]
\centering
\small
\begin{tabular}{rrrrrr}
\hline
$F$ & Aged & Fresh & Adjusted & Donor target & Continuation \\
\hline
5  & 0.038 & 0.650 & 0.698 & 0.099 & 0.617 \\
8  & 0.015 & 0.614 & 0.636 & 0.040 & 0.484 \\
\hline
11 & 0.005 & 0.551 & 0.666 & 0.071 & 0.444 \\
12 & 0.016 & 0.336 & 0.555 & 0.056 & 0.354 \\
14 & 0.009 & 0.111 & 0.217 & 0.021 & 0.084 \\
16 & 0.029 & 0.057 & 0.115 & 0.024 & 0.048 \\
\hline
\end{tabular}
\caption{Reuse across hop indices with the same source-$2$ donor.
Aged, Fresh, and Adjusted report the recipient-relation target at
$F+1$. The final two columns report the donor target at $F+1$ and
the counterfactual continuation at $F+2$, both under the adjusted
intervention. All entries are separately maximized batch-mean
probabilities over loops. Rows below the divider evaluate the first
counterfactual hop beyond the training depth of 11.}
\label{tab:kg-phase-transfer}
\end{table}

The adjusted donor supports the receiving position's relation at every
tested frontier, including unseen hop indices, and matches or exceeds
the fresh donor's score. At $F=12$, the recipient target receives
probability $0.555$, compared with $0.056$ for the donor target.
This supports reuse of an early entity representation under a later
relation. The two targets can coincide in individual examples; such
collisions are not filtered. Continuation remains substantial at
$F=12$ ($0.354$) and declines at greater depths, delimiting the range
of effective reuse in these runs.

\paragraph{Entity decodability and computational phase.}
The aged source-$2$ donor still decodes as $e_2$ with mean probability
$0.886$ and argmax accuracy $96.9\%$.
Nevertheless, in the 11-hop setting at $F=5$, its next-hop score is
$0.038$, compared with $0.650$ for the fresh donor.
Adding $v_2$ raises the aged score to $0.698$; subtracting $v_2$
from the fresh donor reduces its score to $0.031$.
The same direction also transfers between source positions:
adding $v_2$ to an aged source-$6$ donor at $F=14$ raises its score
from $0.006$ to $0.377$.
These controls show that entity decodability alone does not determine
a representation's usefulness for composition, and support a shared
residual-state property affecting that usefulness.
We refer to this property as computational phase. Fresh and aged
states also differ in norm and geometry, so this term does not identify
a separate, isolated gate.

\paragraph{The receiving position must also be prepared.}
We retain a fresh donor at $F$ and replace only the receiving
position's residual before the first layer of the intervention pass.
Using the same receiver's loop-0 residual, from before any recurrent
processing, reduces the next-hop score from $0.650$ to $0.057$ at
$F=5$ in 11-hop queries, and from $0.222$ to $0.015$ at $F=14$
in 29-hop queries. These cases use source positions 2 and 6,
respectively. The loop-0 residual decodes the correct relation with
probability approximately 1.
In contrast, using that receiver's residual from the beginning of
the frontier's emergence loop supports scores of $0.645$ and $0.788$.
Thus relation-token identity alone is insufficient: processing prepares
the receiving residual for composition, and this preparation can precede
the frontier's threshold crossing.

\paragraph{Conclusion: reuse of composition across hop indices.}
Together, these interventions provide causal evidence that DR-Loop
can reuse an intermediate entity representation in a later
composition step, including beyond the training depth.
The same source-$2$ donor and fixed adjustment support predictions
that follow the relation at the receiving position across multiple
frontiers. At $F=12$, for example, the recipient-relation target
has a peak batch-mean probability of $0.555$, compared with $0.056$
for the donor's original next entity. The corresponding continuation
at $F+2$ provides additional evidence that the intervention can
support further counterfactual computation.

Successful reuse nevertheless depends on the form of the residual
representations. An aged donor can retain a decodable entity while
being ineffective for the next composition step; a shared direction
can restore that effectiveness across source positions and receiving
frontiers. Likewise, a receiving residual that clearly encodes the
correct relation can fail to support composition until it has
undergone recurrent processing. These findings support a mechanism
in which recurrence prepares compatible operand and receiver states,
allowing a shared composition computation to operate as the frontier
advances.

This provides mechanistic support for the main paper's account of
length generalization through reuse of a local composition rule.
The declining transfer and continuation scores at later frontiers
show that the demonstrated reuse has a finite effective range.
The results support functional reuse across hop indices, without
establishing exact positional invariance or an exclusively local
attention pathway.


\section{Extreme-Length Failure Analysis}
\label[appendix]{app:extreme_length_failure}

\paragraph{Evaluation scope and conventions.}
We analyze the knowledge-graph checkpoints used in \Cref{sec:RQ1}, each trained on inputs containing at most 11 relation applications. Both use three shared layers, two attention heads per layer, and no positional embeddings. We write an input as a starting entity followed by $K$ relations, and let $n=K+1$ count these tokens. MR-Loop additionally appends \texttt{[EQ]}. Unless otherwise stated, loop indices $t$ are zero-based. MR-Loop's intermediate after loop $t$ is evaluated against $e_{t+1}$; its prescribed final answer is evaluated after $R=n$ passes, at $t=n-1$. DR-Loop stores intermediate $e_k$ at token position $k$ and reads the answer at \texttt{PAD}. Its intermediate readouts use final normalization and unembedding applied to the residual after the last shared layer. DR-Loop's analyzed checkpoint has no input injection; MR-Loop reinjects the original input embeddings at every pass. The common failure account therefore does not depend on both architectures omitting input injection.

The following cohorts and metrics are kept separate. Confidence scores are softmax probabilities from the relevant readout, not binary correctness rates. Multiple hops within a trajectory are not independent samples. These analyses establish properties of the evaluated checkpoints, examples, and recurrence budgets.

\subsection{Behavioral Deterioration and MR-Loop State Localization}
\label{app:mr_extreme_length}

\paragraph{Length and recurrence budgets.}
On the matched-prefix cohort of 64 walks, MR-Loop's final-answer accuracy at $R=n$ is $100\%$ at $n=12$, $70.3\%$ at $n=20$, and $1.6\%$ at $n=30$. A separate natural-walk cohort of 32 examples per length gives $25\%$, $0\%$, and $0\%$ at $n=24,32,48$. These estimates are distinct from the recovery cohort below. For DR-Loop, 64 sampled 29-hop walks and their 11- and 19-hop prefixes give best final-answer accuracies of $0.969$, $0.406$, and $0.031$ at 11, 19, and 29 hops, respectively. The corresponding best budgets are 9, 15, and 14 passes. These budgets are selected retrospectively on the evaluation examples to diagnose whether additional recurrence restores performance. On 19-hop inputs, accuracy instead declines to $0.234$ at 30 passes. Thus the two architectures' headline accuracies use different budget conventions and are not a controlled comparison of model quality.


\subsubsection{Degradation of the [EQ] state limits length generalization} 
\label[appendix]{app:mr_loop_RQ2_eq} 
\paragraph{Testing whether the incoming \texttt{[EQ]} state explains failure on longer inputs.}
We ask whether a transition that fails in a long prompt can be restored by supplying the \texttt{[EQ]} state produced by a shorter prompt requiring the same transition. We construct 64 pairs of inputs:

$$
\begin{aligned}
x_{\mathrm{short}}
&=
[e_0,r_1,\ldots,r_{19},\texttt{[EQ]}],\\
x_{\mathrm{long}}
&=
[e_0,r_1,\ldots,r_{19},r_{20},\ldots,r_{29},\texttt{[EQ]}].
\end{aligned}
$$

These are length-20 and length-30 inputs under our convention that \(n\) counts tokens before \texttt{[EQ]}. Each pair shares the starting entity and first 19 relations, so the correct intermediate entities \(e_1,\ldots,e_{19}\) are identical. Causal attention prevents those shared input positions from seeing the extra relations in the longer prompt, whereas \texttt{[EQ]} can attend to the entire prompt. Because causal attention prevents earlier tokens from accessing the appended relations, the shared relation tokens evolve identically in the short and long runs. We verify this by comparing their hidden vectors at the same loops, both before and after the shared transformer stack; the vectors agree to numerical precision. 

\paragraph{Intervention and readout.}
For each hop \(k\), we run both inputs normally up to the beginning of loop \(t=k-1\), whose intended transition is

$$
e_k = T(e_{k-1},r_k).
$$

Immediately before block 0, we replace the long run's entire \texttt{[EQ]} residual with the short run's \texttt{[EQ]} residual from that same loop. All other positions retain their long-run activations. We then complete that loop and check whether the output argmax at \texttt{[EQ]} equals \(e_k\). For example, testing hop 14 means transplanting the state immediately before loop 13 and scoring \(e_{14}\) at the end of that loop. Each tested hop starts from a separate, otherwise unmodified run; transplants at different hops are not accumulated into one trajectory.

\paragraph{Results.}
Over hops \(12\text{--}19\), mean intermediate-entity accuracy increases from \(5.5\%\) in the unmodified long runs to \(83.6\%\) after transplantation, close to the short runs' \(84.0\%\). These averages cover eight hops on each of 64 paired walks. Mean attention from \texttt{[EQ]} to the required relation \(r_k\) in head L1H1 also increases from \(0.06\) to \(0.43\). This attention change emerges from the residual transplant: attention weights are not directly edited, and the recipient's relation-key states remain unchanged under causal attention. Together, the result show that longer prompts degrade the computational usability of the recurrent [EQ] state, making it a major contributor to extreme-length failure.

\subsubsection{Progress-cue degradation}
\label[appendix]{app:mr_loop_RQ2_procue}

\textbf{Experimental setup.}
We ask whether the relation-state distinctions that guide traversal at early hops remain recognizable at later hops. We analyze 64 length-30 walks and record the residual at each relation position \(k\), immediately before the first Transformer block. Position \(k\) contains the \(k\)th relation, whose scheduled use occurs at zero-based loop \(t=k-1\). For each example, we collect three snapshots of that same relation token: \emph{unread} at \(t=k-5\), four loops before scheduled use; \emph{live} at \(t=k-1\), during scheduled use; and \emph{consumed} at \(t=k+1\), two loops afterward. These labels follow the intended traversal schedule; they do not assume that a long run actually selects the relation or executes its transition correctly.

\textbf{Separation along the early-hop direction.}
Using positions \(k=5,\ldots,11\), which lie within the model's training hop range, we average the paired live-minus-consumed residual differences across positions and examples, then normalize the resulting vector to a unit direction \(v\). Holding \(v\) fixed, we measure the average projection of the live-minus-consumed difference at each later position onto this direction. This measures how strongly the distinction aligns with the early-hop cue; it is expressed in residual-space units, rather than as an accuracy or probability. The projection decreases from \(28.0\) at \(k=11\) to \(9.1\) at \(k=23\), indicating weaker separation along the same direction at greater depth.

\textbf{Transfer of status classifiers.}
We separately construct two linear probes: live versus unread and live versus consumed. Each probe uses the normalized difference between its class means at positions \(5\)--\(11\), with a classification threshold halfway between their mean projections. Keeping both direction and threshold fixed, we evaluate later positions in the same walks. Balanced accuracy---the average of the two class-specific accuracies---is approximately \(99\%\) for both probes at positions \(13\)--\(18\). At positions \(19\)--\(25\), it falls to \(49.8\%\) for live versus unread and \(73.1\%\) for live versus consumed. The latter combines \(46.2\%\) accuracy on live snapshots with \(100\%\) on consumed snapshots, showing that deterioration is asymmetric across statuses. This experiment tests transfer across positions, rather than across independently held-out walks. Together, the reduced separation along the early-hop direction and the lower transfer accuracy of the fixed status probes indicate that the progress cue degrades at greater depth.


\subsection{DR-Loop Intermediate Quality Decreases with Depth}
\label{app:dr_loop_RQ2}

\paragraph{Evaluation setup.}
We examine the knowledge-graph checkpoint trained on at most 11 relation
applications, using 512 sampled 29-hop walks evaluated for 30 recurrent
passes. Let $e_k$ be the correct entity after $k$ relation applications, and let $p_{b,t,k}$ denote the logit-lens probability of the correct
intermediate entity $e_k$ at position $k$ in example $b$ after zero-based loop $t$. 

We distinguish upstream confidence from downstream continuation to assess
whether poor successor representations are associated with weak preceding
intermediates. For a selected upstream hop $F$, we examine confidence in
$e_F$ and continuation to $e_{F+1}$: for example, $F=16$ examines the
transition from $e_{16}$ to $e_{17}$. We consider $F=5$ within the
training range and $F\in\{12,14,16\}$ beyond it. 

We choose the reference loops using an initial 64-example subset.
After each loop, we compute the average probability of the correct
entity at each hop across these examples. We identify the largest hop
index whose average probability is at least $0.5$, using it to estimate
how far the computation has progressed. When this index equals a
selected hop $F$, we use the following loop as its reference time $t_F$. This gives $(F,t_F)\in\{(5,3),(12,8),(14,10),(16,12)\}$.
For example, after loop 11, hop 16 is the deepest hop meeting the
confidence threshold, so we set $t_{16}=12$ to examine continuation
toward hop 17. These reference times are held fixed across all
512 examples.

 We define upstream confidence as the highest probability
assigned to $e_F$ up to and including loop $t_F$, and successor confidence
as the highest probability assigned to $e_{F+1}$ from that loop onward:
\[
u_{b,F}=\max_{0\leq t\leq t_F}p_{b,t,F},
\qquad
v_{b,F}=\max_{t_F\leq t<30}p_{b,t,F+1}.
\]
Taking the maximum separately for each example accommodates differences
in when confidence peaks; the resulting averages measure intermediate
confidence rather than final-answer accuracy. We compare the mean successor score across all examples with the mean
among examples satisfying $u_{b,F}\geq0.8$. This asks whether continuation
improves when the upstream entity was confidently decodable at least once.

\paragraph{Intermediate representations deteriorate with depth.}
The mean upstream score $\mathbb{E}_b[u_{b,F}]$ decreases from $0.982$
at $F=5$ to $0.927$, $0.864$, and $0.713$ at $F=12,14,16$,
respectively. The number of examples reaching $u_{b,F}\geq0.8$ falls
from $503/512$ to $472/512$, $435/512$, and $349/512$ across these
positions. Thus, later intermediates are less consistently decodable
with high confidence, even when each example is evaluated at its own
best observed loop within the specified window. Further recurrence can
also weaken previously computed intermediates: in the initial 64-example
subset, mean correct-entity probability over the same early positions
$k=1,\ldots,11$ decreases from $0.769$ at loop 10 to $0.540$ at loop 29.
These observations establish deterioration of the native intermediate
readouts across composition depth and during additional recurrence. 

\paragraph{Intermediate confidence and downstream continuation.}
At $F=16$, the mean successor score $\mathbb{E}_b[v_{b,F}]$ is
$0.547$ (95\% bootstrap interval $[0.508,0.586]$) across all examples,
compared with $0.776$ ($[0.738,0.811]$) among the 349 examples with
$u_{b,F}\geq0.8$. The corresponding conditional scores at
$F=5,12,14$ are $1.000$, $0.968$, and $0.893$.
Intervals use 2,000 percentile bootstrap resamples of examples. Examples with confidently decodable upstream intermediates therefore
have higher mean successor confidence than the full cohort, although
continuation remains weaker at larger hop indices even after this
filtering. Together with the observed decline in intermediate confidence
with depth, these results are consistent with the hypothesis that
degradation of intermediate representations contributes to length
generalization failure at extreme lengths. 



\paragraph{A brief disruption can persist through subsequent updates.}
In the initial 64-example subset, we transplant the residual at position
13 from the end of loop 8 to that position before block 0 of the same
loop, exposing the computation to this post-update residual one pass
early. Only one position and loop are patched. Over loops $8,\ldots,29$,
the maximum batch-mean probability of $e_{14}$ falls from $0.777$ to
$0.040$, and that of $e_{15}$ falls from $0.723$ to $0.067$.
The same type of perturbation also damages
11-hop inputs, reducing the corresponding score for $e_6$ from
$0.965$ to $0.201$; it reveals a general vulnerability to persistent
disruption rather than an effect unique to extreme lengths.
Together with the native deterioration above, this supports declining
state reliability and limited correction of induced errors. These interventions provide causal evidence that downstream computation
depends on the intermediate residual state, and that a perturbation
confined to one position and loop can cause disruption that persists
within the evaluated loop budget.

\subsection{Computational Status and the Interpretation of Recovery}
\label{app:computational_status_recovery}

\Cref{sec:computation_status} and its supporting experiments show that semantic identity and computational usability can dissociate. In DR-Loop, an aged source entity retains $96.9\%$ argmax decodability but its next-hop score falls from $0.650$ for a fresh donor to $0.038$. Adding the shared fresh--aged direction increases the score to $0.698$, while subtracting it from a fresh donor reduces the score to $0.031$. The receiving residual also matters: substituting its loop-0 state reduces the score from $0.650$ to $0.057$ in the short setting despite retaining the relation identity. In MR-Loop, replacing a live relation with its consumed residual disrupts selection and the transition, while adding the live--consumed direction restores them in the tested settings, including beyond training. Together, these results suggest that correcting a failed hop through the same circuit requires both correct semantic content and a compatible computational status in the states needed to recompute it. Earlier intermediate values must remain usable as operands, and the relevant relation and receiving state must be prepared for the intended transition. Recovery therefore imposes an additional requirement beyond retaining or restoring the correct values: the model must also preserve or reestablish the computational conditions under which those values can be used. 


\subsection{Native Recovery After the First MR-Loop Error}
\label{app:mr_firsthopfailure_recovery}

\paragraph{Cohort and scoring.}
We evaluate 64 independently sampled natural knowledge-graph walks at
each of $n=20$ and $n=30$, using the MR-Loop checkpoint described in
\Cref{sec:mr_loop_q1}. These are unmodified forward passes, with no
activation intervention. The two lengths are sampled separately,
rather than constructed as prefixes of the same walks, and form
separate cohorts from the matched-prefix experiment. Here, $n$ counts
the initial entity and relation tokens before \texttt{[EQ]}, so each
input contains $n-1$ relation applications:
\[
e_0 \xrightarrow{r_1} e_1 \xrightarrow{r_2} \cdots
\xrightarrow{r_{n-1}} e_{n-1},
\qquad e_k=T(e_{k-1},r_k).
\]
The entities $e_k$ are the ground-truth states obtained by executing
the relations on the graph.

\textit{Loop-to-hop alignment.}
We run $R=n$ recurrent passes, indexed by $t=0,\ldots,n-1$, and let
$\hat e_t$ denote the argmax prediction at \texttt{[EQ]} after pass
$t$. Under the intermediate-readout schedule used for this checkpoint,
$\hat e_t$ is compared with $e_{t+1}$ for $0\leq t\leq n-2$.
Thus, the first prediction $\hat e_0$ is evaluated against $e_1$,
and $\hat e_{n-2}$ against the last entity $e_{n-1}$.
We separately evaluate the prescribed final answer after all $R=n$
passes, comparing $\hat e_{n-1}$ with the same target $e_{n-1}$.
This additional pass has no new graph-transition target.
Separating these readouts allows the last intermediate and the
prescribed final answer to differ in correctness.

\textit{Locating the first error.}
To study what happens after an intermediate mistake, we define
\[
t^\star
=
\min\left\{
t\in\{0,\ldots,n-3\}:
\hat e_t\neq e_{t+1}
\right\}.
\]
Using the first error ensures that all earlier scheduled intermediate
predictions were correct. We stop the search at $n-3$, one pass before
the last intermediate readout, to leave at least one subsequent
intermediate on which recovery can be assessed. For example, at
$n=20$, an error at $t^\star=17$ leaves the intermediate readout at
$t=18$ and the separate final-answer readout at $t=19$.
If the search set is empty, $t^\star$ is undefined and the example is
excluded from the post-error analyses. Such an example may still fail
at the last intermediate or final-answer readout.

\textit{Post-error outcomes.}
Among examples with a defined $t^\star$, we measure:
\begin{enumerate}
    \item \textbf{Any later correct intermediate.}
    Whether there exists $u\in\{t^\star+1,\ldots,n-2\}$ such that
    $\hat e_u=e_{u+1}$. This detects a return to the true walk,
    including a transient one.

    \item \textbf{Persistence after the first return
    (reported as sustained recovery).}
    If a later correct intermediate exists, let $u^\star$ be its
    earliest occurrence. We require $\hat e_u=e_{u+1}$ for every
    $u=u^\star,\ldots,n-2$. A return followed by another error fails
    this criterion, even if the trajectory subsequently returns again.
    No minimum recovered suffix length is imposed; a first return at
    $u^\star=n-2$ satisfies the criterion.

    \item \textbf{Final-answer correctness after an error.}
    Whether $\hat e_{n-1}=e_{n-1}$. This measures success at the
    prescribed budget despite an earlier mistake, without requiring
    a fully correct intermediate trajectory.

    \item \textbf{Wrong-state continuation.}
    Whether the next native prediction satisfies
    \[
    \hat e_{t^\star+1}
    =
    T\!\left(\hat e_{t^\star},r_{t^\star+2}\right).
    \]
\end{enumerate}

The last equality is a hypothesis tested against the model's output,
not the definition of $\hat e_{t^\star+1}$.
At the error, the model predicts $\hat e_{t^\star}$ instead of the
true entity $e_{t^\star+1}$. The next scheduled relation is
$r_{t^\star+2}$, so applying it to the erroneous prediction gives the
counterfactual next entity on the right-hand side.
Agreement asks whether the immediately following prediction is
consistent with continuing from that wrong entity; returning to the
true walk instead means $\hat e_{t^\star+1}=e_{t^\star+2}$.
We inspect this next pass because it is the earliest opportunity to
distinguish these behaviors. The model continues from its native
residual throughout; we do not feed the decoded wrong entity back
into it.

\paragraph{Results at $n=20$.}
There are $21/64$ first-error trajectories, with mean first-error loop $12.24$. Final-answer accuracy over all examples is $42/64=65.6\%$. Among first-error trajectories, $2/21$ have a later correct intermediate, $0/21$ satisfy sustained recovery, $1/21$ has a correct prescribed final answer, and $0/21$ follows the exact next transition from its erroneous argmax.

\paragraph{Results at $n=30$.}
All $64/64$ trajectories have a first error, with mean first-error loop $6.75$. Final-answer accuracy is $0/64$. Later correct intermediate hits occur in $22/64$ trajectories, but sustained recovery and final-answer correctness after an error are both $0/64$. Exact wrong-state continuation occurs in $2/64$ cases. The mean argmax probability at the first error is approximately $0.035$, consistent with a diffuse readout rather than a strongly represented alternative entity.

\paragraph{Interpretation: transient returns without reliable repair.}
After the first intermediate decode error, MR-Loop rarely restores
correct subsequent computation in these cohorts. Some trajectories
return to the correct entity at later hops, but these returns do not
persist through the last scheduled intermediate. They therefore
demonstrate intermittent agreement with the true walk, rather than
reliable recovery of the remaining trajectory. The single correct
prescribed final answer at $n=20$ shows that success after an earlier
error is possible, but it is uncommon; no such success is observed at
$n=30$. Together, these results are consistent with the learned
recurrence lacking an effective endogenous procedure for repairing
a failed update within the tested horizons.

\section{Task-Level Compressibility as an Interpretive Control}
\label[appendix]{app:compressibility}

\paragraph{Attribution and purpose.}
~\citet{liang2026latent} introduce a task-level
compressibility framework for sequential reasoning, showing that
many-to-one transitions can irreversibly merge distinctions between
intermediate states, whereas bijective transitions preserve them.
Their analysis includes the prime--composite distinction for polynomial
iteration and its generalization to finite-state function composition
(FSC) through transition rank.

We use this framework here only as an \emph{interpretive control} for
our looped-Transformer experiments. The underlying compressibility
results are not contributions of the present work. We ask whether task-side
compressibility helps organize two empirical phenomena: when recurrent
length generalization is successfully learned, and, conditional on
success, whether the learned computation recovers the intended
intermediate trajectory. 

\subsection{A Task-Side Notion of Compression}

Consider a finite state space $\mathcal{S}$ and a sequence of operators
$o_1,\ldots,o_H$, with
\[
    s_t = T_{o_t}(s_{t-1}),
    \qquad t=1,\ldots,H.
\]
The target is the final state $s_H$. 
For a fixed suffix $o_{t+1:H}$, define
\[
    G_t
    :=
    T_{o_H}
    \circ T_{o_{H-1}}
    \circ \cdots
    \circ T_{o_{t+1}}.
\]
For $t=H$, we take
$G_H=\operatorname{id}_{\mathcal S}$.

Two possible states $u,v\in\mathcal{S}$ at step $t$ are
indistinguishable with respect to the final answer whenever
\[
    G_t(u)=G_t(v).
\]
The number of distinctions that remain relevant after step $t$ can
therefore be summarized by
\[
    \rho_t := |\operatorname{Im}(G_t)|.
\]

When $\rho_t < |\mathcal{S}|$, the suffix merges some possible
intermediate states. In this case, exact recovery of $s_t$ contains
more information than is required for final-answer prediction: a
coarser representation that preserves only the suffix-induced
equivalence class can in principle suffice. We refer to such dynamics
as \emph{compressive}. When $\rho_t=|\mathcal{S}|$, the suffix is
injective and preserves all state distinctions; we refer to this
setting as \emph{non-compressive}.

This distinction is purely a property of the task dynamics. It does
not imply that a neural model must represent the literal symbolic
state $s_t$, nor that $s_t$ must be linearly decodable from a single
residual position. A model may use an invertible reparameterization,
a distributed representation, or another sufficient statistic. The
distinction only asks whether the task itself permits different
running states to be irreversibly merged without changing the final
answer.

\subsection{Instantiation in Our Task Families}

Our three task families provide controlled examples along this axis.

\paragraph{Polynomial iteration.}
For
\[
    s_t = s_{t-1}x_t + 1 \pmod M,
\]
~\citet{liang2026latent} show that the update is
bijective exactly when $x_t$ is a unit modulo $M$. Thus, with a prime
modulus and nonzero multipliers, every update is bijective and the
dynamics contain no task-induced contraction. With a composite
modulus, non-unit multipliers can induce many-to-one updates that
merge possible running states. We therefore use prime and composite
moduli as controlled non-compressive and potentially compressive
settings, respectively.

\paragraph{Finite-state function composition.}
For FSC,
\[
    s_t = f_{\sigma_t}(s_{t-1}),
\]
the corresponding task-side quantity is the transition rank
$|\operatorname{Im}(f_i)|$. A bijective function preserves all state
distinctions, whereas a lower-rank function merges some possible
incoming states. Fully bijective FSC therefore provides an
information-preserving control, while libraries containing
non-bijective functions permit task-induced compression
\citep{liang2026latent}.

\paragraph{Knowledge-graph traversal.}
In our knowledge-graph construction, relation maps are permutations
over entities. The resulting transitions are therefore bijective and
do not provide the same many-to-one shortcut structure. This makes the KG task a complementary information-preserving setting for examining whether the recurrent models maintain the intended
intermediate entity trajectory.

\subsection{Implications for Learnability and Intermediate-State Faithfulness}

The task-side distinction above helps organize two patterns in our
experiments. First, settings that admit compression often support
behavioral length generalization more readily. The clearest example is
polynomial iteration: both architectures succeed substantially more
often on the tested composite moduli than on the tested prime moduli.
We interpret this as consistent with compressive tasks admitting a
larger class of sufficient solutions, including coarser representations
that need not preserve distinctions already erased by the remaining
computation. Information-preserving tasks do not offer this particular
shortcut and therefore impose a stronger information-preservation
burden on any successful solution. This is an interpretation of the
observed learnability differences, not a claim that compression
guarantees easier optimization.

Second, compressibility helps explain the separation between behavioral
success and intermediate-state faithfulness. In several successful
composite-polynomial and partially compressive FSC settings, the models
recover the final answer despite weak recovery of earlier ground-truth
intermediates. Such behavior is compatible with the task dynamics
because those earlier distinctions need not all remain relevant to the
final answer. By contrast, successful MR-Loop models in our
information-preserving settings---including KG, fully bijective FSC,
and polynomial $m=11$---tend to expose substantially more complete
task-aligned trajectories.

Compressibility nevertheless does not determine either outcome.
Some compressive settings still fail to extrapolate or learn nearly
faithful trajectories, while some information-preserving settings fail
to learn a successful length-generalizing solution. Model capacity,
optimization, recurrence allocation, and task difficulty also matter.
We therefore use compressibility as a constraint on the available
solution space: it helps explain both why some tasks admit easier
length-generalizing strategies and why successful prediction can, in
some settings, be separated from recovery of the full ground-truth
trajectory. 

\section{Additional Results for RQ3}
\label[appendix]{app:RQ3_additional_results} 

\subsection{Behavioral Length Generalization Across Task Configurations}
\label[appendix]{app:RQ3_additional_length_tasks}

Table~\ref{tab:generalization_summary} reports whether MR-Loop and DR-Loop
successfully length-generalize across the polynomial, finite-state
function composition (FSC), and knowledge-graph (KG) task families.
This sweep tests how robustly recurrent length extrapolation emerges as
we vary task structure and difficulty.

Length generalization is strongly task- and model-dependent. On
polynomial iteration, MR-Loop generalizes for all tested composite
moduli and for $m=11$, but not for the remaining tested prime moduli.
DR-Loop similarly fails on the tested prime-modulus settings and
succeeds on several, but not all, composite-modulus settings. On FSC,
MR-Loop succeeds across all tested configurations, whereas DR-Loop
succeeds when the transition library contains $8$ functions but fails
in the corresponding $16$-function settings across the state-space
sizes and bijective fractions considered. On KG traversal, MR-Loop
successfully extrapolates in all six tested configurations, while
DR-Loop succeeds in three.

These results show that weight sharing and additional recurrent
computation do not by themselves guarantee length generalization.
Whether extrapolation emerges depends jointly on the task configuration
and the learned recurrent system. Across the configurations considered
here, MR-Loop exhibits substantially more robust behavioral
extrapolation than DR-Loop. Section~\ref{sec:RQ3} complements this
behavioral comparison by asking a separate question: when
length generalization succeeds, does the recurrent computation also
recover the intended intermediate trajectory?

\begin{table*}[htbp]
\centering
\caption{Length-generalization success across task configurations.
$\checkmark$ indicates successful generalization and $\times$ indicates failure.}
\label{tab:generalization_summary}

\small
\begin{tabular}{lcccccccc}
\toprule

\multicolumn{9}{c}{\textbf{Polynomial}} \\
\midrule
Modulus
    & 11 & 19 & 25 & 29 & 30 & 35 & 40 & 50 \\
MR-Loop
    & $\checkmark$ & $\times$ & $\checkmark$ & $\times$
    & $\checkmark$ & $\checkmark$ & $\checkmark$ & $\checkmark$ \\
DR-Loop
    & $\times$ & $\times$ & $\checkmark$ & $\times$
    & $\checkmark$ & $\checkmark$ & $\checkmark$ & $\times$ \\

\midrule
\multicolumn{9}{c}{\textbf{Finite-State Composition (FSC)}} \\
\midrule
$\alpha$
    & \multicolumn{4}{c}{0.5}
    & \multicolumn{4}{c}{1.0} \\
$N_{\text{state}}$
    & \multicolumn{2}{c}{16}
    & \multicolumn{2}{c}{32}
    & \multicolumn{2}{c}{16}
    & \multicolumn{2}{c}{32} \\
$N_{\text{function}}$
    & 8 & 16 & 8 & 16 & 8 & 16 & 8 & 16 \\
MR-Loop
    & $\checkmark$ & $\checkmark$ & $\checkmark$ & $\checkmark$
    & $\checkmark$ & $\checkmark$ & $\checkmark$ & $\checkmark$ \\
DR-Loop
    & $\checkmark$ & $\times$ & $\checkmark$ & $\times$
    & $\checkmark$ & $\times$ & $\checkmark$ & $\times$ \\

\midrule
\multicolumn{9}{c}{\textbf{Knowledge Graph}} \\
\midrule
$N_{\text{entities}}$
    & \multicolumn{6}{c}{100}
    & \multicolumn{2}{c}{--} \\
$N_{\text{relations}}$
    & 10 & 10 & 20 & 20 & 8 & 12 & -- & -- \\
Degree
    & 5 & 10 & 5 & 10 & 8 & 12 & -- & -- \\
MR-Loop
    & $\checkmark$ & $\checkmark$ & $\checkmark$ & $\checkmark$
    & $\checkmark$ & $\checkmark$ & -- & -- \\
DR-Loop
    & $\times$ & $\checkmark$ & $\times$ & $\times$
    & $\checkmark$ & $\checkmark$ & -- & -- \\

\bottomrule
\end{tabular}
\end{table*}

\subsection{Intermediate-State Recovery Across Tasks}
\label{app:rq3-intermediate}

We complement the behavioral length-generalization results in
\Cref{app:RQ3_additional_length_tasks} by examining whether successful models recover
the intended ground-truth intermediate states during recurrent
computation. We use the logit-lens procedure described in
\Cref{app:logit-lens}. These analyses characterize intermediate-state
decodability and should therefore be interpreted as representational,
rather than by themselves causal, evidence.

\begin{figure}[htbp]
    \centering
    \includegraphics[width=1\linewidth]{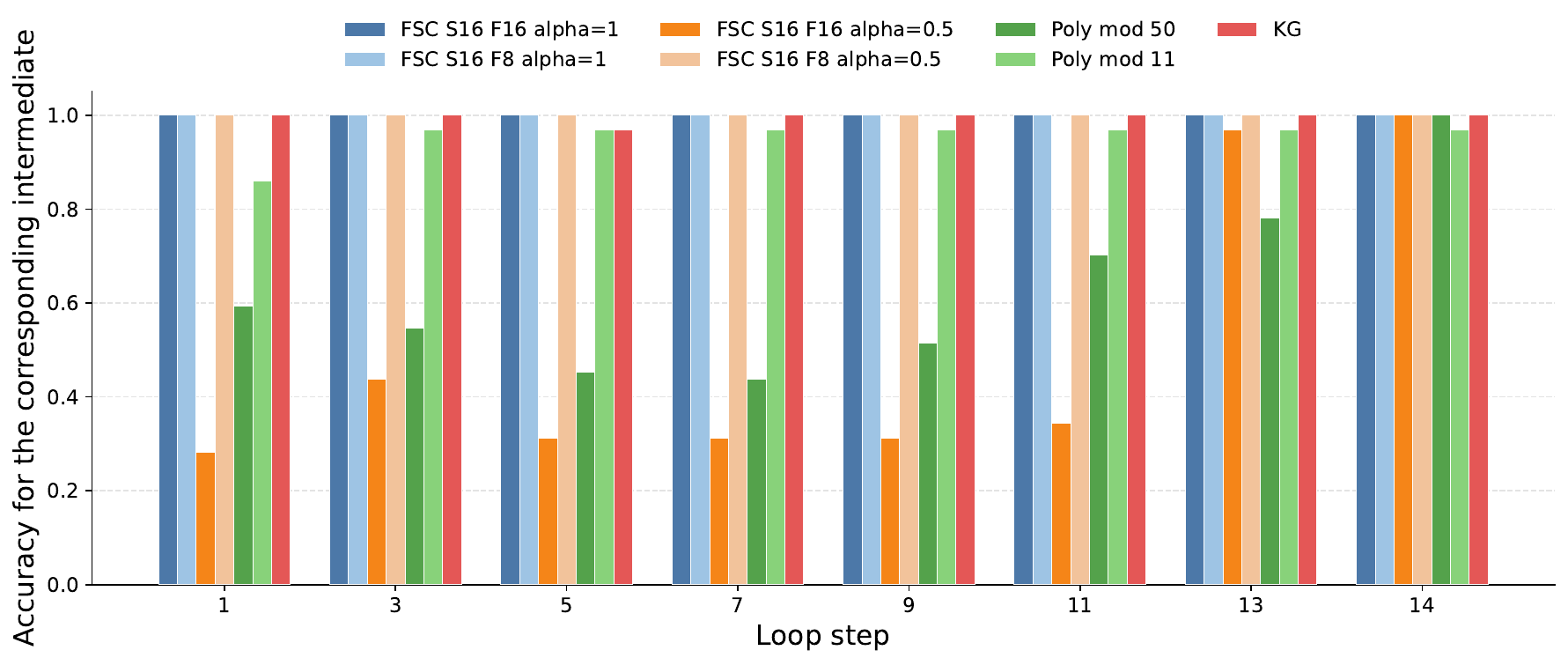}
    \caption{\textbf{Accuracy of ground-truth intermediate values at the corresponding recurrent steps across tasks for MR-Loop}. Easier settings and tasks that do not admit shortcut solutions consistently recover the correct intermediate states throughout the computation. In contrast, more difficult or compressible settings, such as polynomial iteration with \(m=50\) and FSC with \(S=16, F=16\), fail to faithfully represent the intermediate trajectory and recover the correct states only near the final steps.}
    \label{fig:mrloop_intermediates_acc}
\end{figure}

\paragraph{MR-Loop.}
\Cref{fig:mrloop_intermediates_acc} compares recovery of the corresponding
ground-truth intermediate across several task configurations. The
quality of the intermediate trajectory varies substantially despite
successful behavioral extrapolation. Knowledge-graph traversal,
polynomial iteration with $m=11$, and the easier FSC configurations
recover the intended state with high accuracy across most of the
recurrent computation. In contrast, polynomial iteration with $m=50$
and the more difficult FSC configuration with $S=16,F=16, \alpha=0.5$ show much
weaker recovery at earlier steps, with the correct intermediate becoming
reliably decodable mainly near the final updates.

Thus, MR-Loop can arrive at the correct extrapolated answer without
uniformly exposing the full intended intermediate trajectory. The
contrast is consistent with the task-side compressibility perspective
in \Cref{app:compressibility}, while also showing that compressibility
alone does not determine the learned representation.

\begin{figure}[htbp]
    \centering
    \includegraphics[width=1\linewidth]{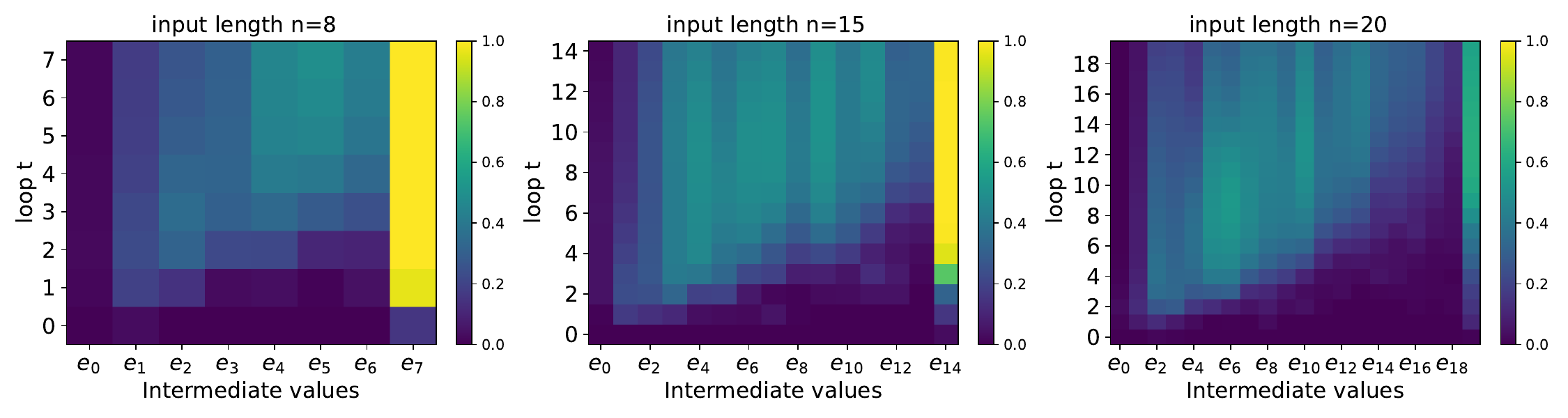}
    \caption{
\textbf{DR-Loop can recover the final answer without faithfully tracking all intermediate states.}
Batch-mean softmax probability assigned to the ground-truth intermediate state at each write site for finite-state composition (FSC) with $S=16$, $F=8$, and bijective-function fraction $\alpha=0.5$, evaluated at task lengths $n\in\{8,15,20\}$. Despite weak recovery of many intermediate states, particularly at early recurrent steps, the probability assigned to the final target becomes high near the final write site. This dissociation shows that successful final-answer prediction does not necessarily imply faithful step-by-step intermediate-state formation.
}
    \label{fig:drloop_intermediates_weight_fsc}
\end{figure}

\paragraph{DR-Loop.}
DR-Loop exhibits an even clearer separation between final-answer success
and intermediate-state recovery. \Cref{fig:drloop_intermediates_weight_fsc} shows a
representative FSC setting with $S=16$, $F=8$, and bijective-function
fraction $\alpha=0.5$, evaluated at input lengths
$n\in\{8,15,20\}$. At each recurrent pass, we measure the batch-mean
logit-lens probability assigned to each ground-truth intermediate at
its corresponding write site.

Across lengths, many intermediate states remain weakly decodable,
particularly during earlier recurrent passes, even though probability
eventually concentrates on the final target near the end of the
computation. This result provides a concrete
example in which successful final-answer prediction does not require a
fully recoverable step-by-step intermediate trajectory.

Taken together, these results reinforce the distinction between
behavioral length generalization and intermediate-state faithfulness:
successful recurrent extrapolation need not expose the complete
ground-truth state trajectory under the model's output readout.

\section{Generality of the Identified Recurrent Mechanisms}
\label[appendix]{app:mechanism-generality}

The main analyses in \Cref{sec:RQ1} characterize recurrent
mechanisms in representative knowledge-graph (KG) checkpoints.  We next
ask which aspects of these mechanisms recur outside those primary
checkpoints.  We consider two complementary forms of replication.
First, for MR-Loop, we analyze separately trained checkpoints on an
additional KG configuration, Polynomial iteration, and finite-state
composition (FSC).  Second, for DR-Loop, we analyze two additional KG
configurations.  These experiments are intended to test recurrence of
the identified \emph{functional organization}, rather than exact
identity of individual heads or residual directions across models.

Across the tested MR-Loop checkpoints, we find a fixed accumulator at
\texttt{[EQ]} coupled to advancing input selection, together with
live--consumed residual-state distinctions that causally affect routing.
Across the additional DR-Loop checkpoints, we again observe an advancing
computational frontier and a dissociation between semantic decodability
and subsequent compositional usability.  The quantitative details below
also reveal checkpoint-specific differences, delimiting the scope of
these generality claims.

\paragraph{Why DR-Loop replication focuses on KG.}
We evaluate cross-checkpoint replication of the DR-Loop mechanism on
additional KG configurations because the Polynomial and FSC settings
examined in Section~\ref{sec:RQ3} do not consistently exhibit clearly
decodable, task-aligned intermediate trajectories. In several tested
settings, DR-Loop either fails to extrapolate or achieves correct final
predictions while many ground-truth intermediate states remain weakly
decodable under our readout analyses. These settings provide complementary
tests of the distinction between behavioral success and intermediate-state
faithfulness. We therefore use additional KG checkpoints to test recurrence
of the specific advancing-frontier mechanism characterized in
\Cref{sec:dr-mechanism}, and interpret these experiments as replication across KG
configurations rather than evidence of a task-universal DR-Loop mechanism. 

\subsection{MR-Loop: Cross-Task and Cross-Checkpoint Replication}
\label{app:mr_cross_task}

\paragraph{Configurations and conventions.}
We analyze three separately trained MR-Loop checkpoints: a KG task with
100 entities and eight relations available per entity, bijective
finite-state composition with 16 states and eight functions, and
Polynomial iteration
\[
    c_k = (c_{k-1}x_k + 1) \bmod 11.
\]
Each model contains three Transformer blocks and two attention heads per
block.  Input length $n$ counts tokens before \texttt{[EQ]}, including
the initial state, and recurrent loops $t$ are zero-indexed.

The logit lens applies the model's final normalization and output
projection to cached residual states.  Let \(c_k\) denote the ground-truth state after \(k\) task transitions, with \(c_0\) the initial state and \(c_{n-1}\) the final target. Recurrent passes are indexed from \(t=0\), and we decode the \([EQ]\) residual after each pass. In the KG checkpoint, this readout aligns with \(c_{t+1}\): after the first pass (\(t=0\)), it already decodes \(c_1\). We therefore evaluate against \(c_{\min(t+1,n-1)}\), where the minimum caps the target at the final state. In the Polynomial and FSC checkpoints, the established intermediate trajectory instead aligns with \(c_t\). For example, after pass \(t=2\), the evaluated state is \(c_3\) for KG but \(c_2\) for Polynomial and FSC. The reported length-15 decoding accuracies cover \(t=2,\ldots,14\) for Polynomial and \(t=1,\ldots,14\) for FSC, excluding their less reliable initial readouts. These checkpoint-specific offsets describe the observed correspondence between recurrent passes and task transitions; all three checkpoints maintain the evolving intermediate at the same fixed \([EQ]\) position.  The length-15 logit-lens
experiments use 32 examples for KG and FSC and 64 examples for
Polynomial.  The live--consumed interventions use 64 examples for KG
and Polynomial and 32 for FSC.

\paragraph{Live--consumed direction and intervention protocol.}
For example $i$, token position $k$, and residual site $s$, define
\[
    \Delta h_{s,i,k}
    =
    h^{\mathrm{live}}_{s,i,k}
    -
    h^{\mathrm{cons}}_{s,i,k},
\]
and
\[
    v_s
    =
    \mathbb{E}_{i,k}
    \left[\|\Delta h_{s,i,k}\|_2\right]
    \frac{
        \mathbb{E}_{i,k}\left[\Delta h_{s,i,k}\right]
    }{
        \left\|
        \mathbb{E}_{i,k}\left[\Delta h_{s,i,k}\right]
        \right\|_2
    }.
\]
Thus, each checkpoint has one separately estimated direction per
residual site.  We intervene jointly at the input to block~0 and after
each of the three blocks.  At the selected token position, the native
residual is replaced either by a later consumed residual, by the
consumed residual plus $v_s$, or by the corresponding control state.
The input token and incoming \texttt{[EQ]} residual remain unchanged.

For KG, live and consumed snapshots are taken at loops $k-1$ and $k+1$,
and directions are estimated over positions $k=5,\ldots,11$.  For
Polynomial, the corresponding snapshots are $k$ and $k+2$, with fitting
over $k=4,\ldots,7$.  For FSC, the executed configuration uses snapshots
$k-1$ and $k+5$, with fitting over $k=5,6,7$.  Directions are therefore
estimated separately within each checkpoint; no residual vector is
transferred numerically between models.

\paragraph{A fixed accumulator with advancing input selection.}
Despite task-specific differences in heads and loop alignment, all three
checkpoints exhibit the same broad organization.  Attention patterns
advance through successive input positions while \texttt{[EQ]} remains
a fixed query that tracks the evolving computation.  Logit-lens
measurements localize the corresponding intermediate value primarily at
\texttt{[EQ]}, with substantially weaker decoding at the inspected input
positions.

On length-15 inputs, the KG checkpoint decodes the relevant look-ahead
state $c_{t+1} $ with $93.8$--$100\%$ accuracy over the tested recurrent steps.
Polynomial decodes $c_t$ with $95.3$--$98.4\%$ accuracy for
$t=2,\ldots,14$, and FSC decodes $c_t$ with $100\%$ accuracy for
$t=1,\ldots,14$.  The shared feature is therefore not an identical
loop-to-hop offset, but a fixed accumulator coupled to advancing operator
selection.

\paragraph{Live--consumed interventions reproduce the routing effect.}
The live--consumed intervention has a consistent qualitative effect
across the three checkpoints.  Table~\ref{tab:mr_cross_task_status}
reports representative later-position interventions.

\begin{table}[t]
\centering
\small
\setlength{\tabcolsep}{4pt}
\begin{tabular}{llrrrrr}
\toprule
Task $(n,t,k)$ & Metric
    & Native
    & Cons.
    & Cons.$+v$
    & Live$-v$
    & Cons.$+r$ \\
\midrule
KG $(20,13,14)$
    & Accuracy (\%)
    & 93.8 & 7.8 & 93.8 & 7.8 & 7.8 \\
KG $(20,13,14)$
    & L2H1 mass
    & .831 & .001 & .869 & .013 & .000 \\
\midrule
Polynomial $(20,12,12)$
    & Accuracy (\%)
    & 92.2 & 17.2 & 90.6 & 17.2 & 17.2 \\
Polynomial $(20,12,12)$
    & L1H1 mass
    & .824 & .004 & .766 & .006 & .009 \\
\midrule
FSC $(20,12,13)$
    & suffix mass
    & .194 & .025 & .177 & .027 & .040 \\
\bottomrule
\end{tabular}
\caption{
Representative live--consumed interventions across separately trained
MR-Loop checkpoints.  ``Cons.'' replaces the target token with its
consumed residual state; ``Cons.$+v$'' adds the separately estimated
live--consumed direction; ``Live$-v$'' subtracts the direction from a
live state; and $r$ denotes the tested norm-matched random direction.
The attention metrics use different aggregation domains across tasks and
should be compared only within each task.
}
\label{tab:mr_cross_task_status}
\end{table}

For the additional KG checkpoint, replacing key $14$ at loop $t=13$
with its consumed state shifts the relevant attention pointer from key
$14$ to key $15$ and reduces accuracy on $c_{14}$ from $93.8\%$ to
$7.8\%$.  Adding the separately estimated live--consumed direction
returns the pointer to key $14$ and restores accuracy to $93.8\%$.

Polynomial exhibits the same coupled routing-and-computation effect.
At $(n,t,k)=(20,12,12)$, consumed-state replacement shifts the L1H1
pointer from key $12$ to key $13$ and reduces $c_{12}$ accuracy from
$92.2\%$ to $17.2\%$.  Adding the live--consumed direction restores
the pointer and raises accuracy to $90.6\%$.  Subtracting the direction
from a live state reproduces the disruption.

FSC exhibits a closely related routing effect.  At
$(n,t,k)=(20,12,13)$, replacing key $13$ with its consumed state moves
both layer-2 pointers from key $13$ to key $14$.  Mean suffix attention
to key $13$ falls from $0.194$ to $0.025$; adding the live--consumed
direction restores the two pointers and raises the mass to $0.177$.
The FSC current-state readout remains correct in these interventions,
however, so these saved measurements establish causal control of routing
but do not establish recovery of the following state transition.

\paragraph{State swaps separate accumulated content from input selection.}
We next ask whether the state accumulated at \texttt{[EQ]} can be
perturbed without equivalently changing operator selection.

In KG, we replace the incoming \texttt{[EQ]} residual with the
same-loop residual from another example while retaining the recipient's
relation-key states.  At $n=20$ and $k=14$, the output equals the
recipient relation applied to the donor entity with $93.7\%$
counterfactual accuracy, while the L2H1 pointer remains at the
recipient's key~14.  The analogous counterfactual accuracy is $90.3\%$
at $k=11$.  Native-target leakage is zero on these counterfactual
subsets.  This provides direct causal evidence that \texttt{[EQ]}
supplies the entity operand while relation-token states determine the
selected operation.

Polynomial and FSC provide complementary evidence through
within-example temporal state swaps. These experiments replace the
incoming \texttt{[EQ]} state with preceding- or following-loop states
from the same example. For Polynomial at $(n,t)=(15,6)$, native-target
accuracy falls from $98.4\%$ to $6.2\%$ under either substitution while
the L1H1 pointer remains at key~6. For FSC at $(15,10)$, native-target
accuracy falls from $100\%$ to $0\%$, while L1H1 remains at key~10 and
the layer-2 pointers remain at key~11. These interventions support a
partial functional separation between accumulated content and input
selection. We did not perform the corresponding cross-example
donor--recipient composition tests for Polynomial or FSC, so whether
these models exhibit the counterfactual composition behavior observed
in KG remains untested.

\paragraph{Complementary causal evidence for routing.}
Additional interventions support causal dependence on the advancing
routing patterns.  In Polynomial, deleting the \texttt{[EQ]}-to-$t$
attention edge across all heads and layers reduces final accuracy from
$98.4\%$ to $1.6\%$ at $n=8$, and from $96.9\%$ to $10.9\%$ at
$n=15$.  At \(n=8\), rerouting the removed attention mass to a neighboring key similarly reduces final accuracy from $98.4\%$ to $3.1\%$. 

In FSC, the tested previous-token intervention reduces final accuracy
from $100\%$ to $31.2\%$ at $n=8$.  This intervention removes the
scheduled edge from query $t+2$ to key $t+1$ wherever the query exists,
including at the \texttt{[EQ]} boundary, and therefore should not be
interpreted as identifying exactly the same necessary edge or head as in
KG.

Polynomial also shows joint dependence on the accumulator and the
selected input-token representation.  At loop~3, patching clean
block-0 outputs into a corrupted run yields $43.8\%$ immediate
clean-target accuracy when patching \texttt{[EQ]} alone and $12.5\%$
when patching position~3 alone, but $96.9\%$ when both are patched.
This supports a joint contribution of the fixed accumulator and the
selected input position to the immediate recurrent update.

\paragraph{Scope of the MR-Loop replication.}
These experiments support recurrence of a common functional organization:
an advancing selection process interacts with an intermediate state
maintained at a fixed \texttt{[EQ]} accumulator, and the selectability
of input states depends on a transferable live--consumed residual
component.  The claim is intentionally weaker than exact circuit
identity.  The participating heads, loop offsets, and estimated
directions vary across checkpoints, and each direction is fitted
separately within its model.

The Polynomial and FSC experiments establish transfer to positions
excluded from direction fitting. 
The experiments
also use one checkpoint per reported configuration and one tested
random-direction seed.  Finally, successful local routing does not imply
indefinite extrapolation: the identified mechanisms themselves degrade
at sufficiently large depth.

\subsection{DR-Loop: Replication Across Knowledge-Graph Configurations}
\label{app:dr_cross_kg}

The main DR-Loop analysis in \Cref{sec:dr-mechanism} identifies three related
properties: an advancing computational frontier, causal propagation of
intermediate residual states, and a dependence of subsequent composition
on whether an entity representation is fresh or aged.  We test whether
these properties recur in two additional independently trained KG
checkpoints.

Both configurations contain 100 entities and use relation-count /
out-degree pairs $(8,8)$ and $(12,12)$; we refer to them as KG-8 and
KG-12.  As in the primary analysis, the models contain three shared
Transformer layers with two heads per layer, width 256, no positional
embeddings, and no recurrent input reinjection.  Training covers at most
11 relation applications.

\paragraph{Advancing attention and intermediate-state frontiers.}
Both checkpoints exhibit advancing attention patterns and
intermediate-state decoding beyond the 11-hop training horizon.  On
14-hop queries, the correct entities $e_{12}$ and $e_{13}$ remain
strongly decodable at their corresponding relation positions:
\[
\begin{array}{c|cc}
 & e_{12} & e_{13} \\
\hline
\text{KG-8}  & 0.897 & 0.887 \\
\text{KG-12} & 0.895 & 0.919
\end{array}
\]
where entries denote peak batch-mean logit-lens probabilities across
recurrent passes.

The detailed trajectories nevertheless differ.  After 15 passes, the
mean correct-entity probability over positions $1$--$13$ is $0.909$ in
KG-8 but $0.067$ in KG-12: high-confidence logit-lens readouts of earlier intermediates persist longer in KG-8 than in KG-12.  Despite this difference, final PAD accuracy is $82.8\%$ and
$95.3\%$, respectively.  Thus, persistent high-confidence logit-lens decodability of all earlier intermediates at their original relation positions is not required for accurate final readout in these checkpoints.

\paragraph{State swaps causally redirect downstream computation.}
We apply the sustained intermediate-state swap used in the primary
DR-Loop analysis, replacing the residual at position~5 using a donor
trajectory and allowing subsequent recurrent computation to continue.
The intervention redirects the final PAD prediction to the
counterfactual donor continuation in $51.6\%$ of KG-8 examples and
$50.0\%$ of KG-12 examples.  These results provide an additional
causal test that the intermediate residual state influences downstream
composition rather than merely correlating with the final prediction.

\paragraph{Fresh--aged dependence replicates across checkpoints.}
We next repeat the computational-phase intervention using source
position~2.  Let $v_2$ denote the within-checkpoint direction obtained
from the difference between fresh and aged source-2 residual states.
For frontier $F=5$, we transplant either the fresh donor, the aged
donor, the adjusted state $h^{\mathrm{aged}}+v_2$, or
$h^{\mathrm{fresh}}-v_2$.  Table~\ref{tab:dr_cross_kg_phase} reports
the resulting next-hop score, defined as the peak batch-mean
logit-lens probability of the recipient-relation counterfactual target.

\begin{table}[t]
\centering
\small
\setlength{\tabcolsep}{5pt}
\begin{tabular}{lrrrrrr}
\toprule
Checkpoint
& Fresh
& Aged
& Aged$+v_2$
& Fresh$-v_2$
& Aged $P(e_2)$
& Aged argmax \\
\midrule
KG-8
& .612 & .050 & .935 & .032 & .999 & 100\% \\
KG-12
& .434 & .008 & .553 & .008 & .233 & 65.6\% \\
\bottomrule
\end{tabular}
\caption{
Fresh–aged interventions in two additional DR-Loop KG checkpoints, evaluated on 64 examples with input length \(n=12\) (11 relation applications) and \(R=12\) recurrent passes. Source-position-2 residuals are transplanted to frontier \(F=5\) at zero-indexed loop \(t^\star=3\). The four intervention columns report the maximum, over loops \(t=3,\ldots,11\), of the batch-mean logit-lens probability of the counterfactual next-hop target at position 6. This maximum is computed separately for each intervention condition; the scores are not final-answer accuracies. Directions are estimated separately within each checkpoint. 
}
\label{tab:dr_cross_kg_phase}
\end{table}

In both checkpoints, aging strongly reduces subsequent composition, and
adding the fresh--aged direction restores a substantial portion of the
next-hop signal.  Conversely, subtracting the same direction from a
fresh donor nearly eliminates the effect.  KG-8 provides a particularly
clear dissociation between semantic content and computational usability:
the aged residual still decodes as $e_2$ with probability $0.999$ and
$100\%$ argmax accuracy, yet its next-hop score falls from $0.612$ to
$0.050$.  Adding $v_2$ raises this score to $0.935$.

KG-12 exhibits the same causal ordering---fresh $>$ aged,
aged$+v_2 >$ aged, and fresh$-v_2 <$ fresh---despite substantially
weaker persistence of the aged semantic representation.  We therefore
interpret the shared result as replication of \emph{phase-dependent
recurrent reuse}, rather than as requiring an aged state to remain
perfectly decodable in every checkpoint.

\paragraph{Transfer across source positions and receiver preparation.}
The phase intervention also transfers across source positions.  To test transfer across source positions, we use 64 examples with input length \(n=30\) (29 relation applications) and \(R=30\) recurrent passes. At zero-indexed loop \(t^\star=10\), we transplant an aged source-6 donor, adjusted using the source-2 direction, to the later unseen frontier \(F=15\). For this experiment, the next-hop score is the maximum, over loops \(t=10,\ldots,29\), of the batch-mean logit-lens probability of the counterfactual target at position 16, computed separately for each condition. Adding the source-2 direction raises this score from \(0.010\) to \(0.240\) in KG-8, compared with \(0.271\) for the corresponding fresh donor. In KG-12, the analogous intervention raises the score from \(0.009\) to \(0.126\), compared with \(0.151\) for the fresh donor.

The receiving residual must also be in an appropriate computational
state.  Resetting the receiver to an early recurrent state while keeping
the fresh donor fixed reduces the next-hop score from $0.612$ to
$0.040$ in KG-8 and from $0.434$ to $0.021$ in KG-12.  Together,
these results reproduce the primary checkpoint's finding that semantic
identity alone is insufficient: successful recurrent reuse depends on
the computational state of both the propagated operand and the receiving
position.

\paragraph{Scope of the DR-Loop replication.}
The additional KG checkpoints reproduce the central qualitative
organization identified in the primary model: an advancing
intermediate-state frontier, causal influence of intermediate residual
states on downstream composition, and a fresh--aged residual distinction
that controls whether a represented entity can participate effectively
in the next update.  At the same time, quantitative differences are
substantial.  KG-8 preserves earlier intermediate representations much
more strongly than KG-12, illustrating that the persistence of the
representational trace is not itself invariant across checkpoints.

As in the MR-Loop experiments above, residual directions are estimated
separately for each checkpoint; these experiments do not demonstrate a
single vector that transfers between independently trained models.
They establish replication across independently trained task
configurations, but not robustness across random training seeds.

\subsection{What Generalizes Across the Experiments?}
\label{app:generality_summary}

Taken together, the replication experiments identify a level of
description that is more stable than individual attention heads or exact
loop-to-hop alignments.

For MR-Loop, the recurring organization is a fixed accumulator at
\texttt{[EQ]} together with advancing input selection.  Operator-token
states encode whether they remain available for selection, and
live--consumed interventions causally manipulate this status.  This
organization appears in an additional KG checkpoint, Polynomial
iteration, and FSC, although the strength of the downstream
compositional effect differs across tasks.

For DR-Loop, recurrent computation instead advances through sequence
positions.  Across multiple KG configurations, intermediate residuals
causally influence later composition, while the usefulness of an entity
representation depends on whether it remains in a fresh,
computation-compatible state.

Thus, the experiments do not support a claim that looped Transformers
learn a single universal circuit.  Rather, they support a broader
representational principle: \emph{semantic content and computational
usability are distinct properties of a recurrent state}.  The two
architectures instantiate this distinction differently---through
live versus consumed operator states in MR-Loop and fresh versus aged
intermediate states in DR-Loop---but both require task-relevant content
to remain in a representation compatible with the next recurrent
transition.

\end{document}